\documentclass[acmtog,authorversion]{acmart}

\pdfoutput=1

\usepackage{graphicx}

\usepackage{animate}
\usepackage{tabularx}
\usepackage{booktabs}
\usepackage{caption}
\usepackage{tikz}
\usepackage{pgfplots}
\usepackage{pgfplotstable}
\usepackage[skins]{tcolorbox}
\usepackage{standalone}
\usepackage{bm}
\usepackage{mleftright}
\usepackage{textcomp}
\usepackage[normalem]{ulem}
\newlength{\itemwidth}

\newcolumntype{Y}{>{\centering\arraybackslash}X}
\newcolumntype{P}[1]{>{\centering\arraybackslash}p{#1}}

\usetikzlibrary{calc}
\usetikzlibrary{tikzmark}
\pgfplotsset{compat=1.14}

\definecolor{619D47}{RGB}{97,157,71}
\definecolor{EE7F0E}{RGB}{238,127,14}
\definecolor{3787CF}{RGB}{55,135,207}
\definecolor{DBDC4A}{RGB}{219,220,74}

\pgfplotscreateplotcyclelist{custompalette}{
    {619D47!50!black,fill=619D47},
    {EE7F0E!50!black,fill=EE7F0E},
    {3787CF!50!black,fill=3787CF},
    {DBDC4A!50!black,fill=DBDC4A}
}

\makeatletter
\let\zeropad\@anim@pad 
\makeatother

\tikzset{
  double arrow/.style args={#1 with #2 and #3}{
    -stealth, line width=#1, #2, postaction={
        draw, -stealth, line width=(#1)/2, shorten <= (#1)/4, shorten >= 2*(#1)/4, #3 
    }
  }
}

\setcopyright{acmlicensed}
\acmJournal{TOG}
\acmYear{2019}
\acmVolume{38}
\acmNumber{6}
\acmArticle{184}
\acmMonth{11}
\acmDOI{10.1145/3355089.3356528}

\setcitestyle{square}

\begin{document}

\title{3D Ken Burns Effect from a Single Image}

\author{Simon Niklaus}
\affiliation{\institution{Portland State University}}
\email{sniklaus@pdx.edu}

\author{Long Mai}
\affiliation{\institution{Adobe Research}}
\email{malong@adobe.com}

\author{Jimei Yang}
\affiliation{\institution{Adobe Research}}
\email{jimyang@adobe.com}

\author{Feng Liu}
\affiliation{\institution{Portland State University}}
\email{fliu@cs.pdx.edu}

\renewcommand{\shortauthors}{Niklaus, Mai, Yang, and Liu}

\begin{abstract}

    {\let\thefootnote\relax\footnote{\url{http://sniklaus.com/kenburns}}\setcounter{footnote}{0}}The Ken Burns effect allows animating still images with a virtual camera scan and zoom. Adding parallax, which results in the 3D Ken Burns effect, enables significantly more compelling results. Creating such effects manually is time-consuming and demands sophisticated editing skills. Existing automatic methods, however, require multiple input images from varying viewpoints. In this paper, we introduce a framework that synthesizes the 3D Ken Burns effect from a single image, supporting both a fully automatic mode and an interactive mode with the user controlling the camera. Our framework first leverages a depth prediction pipeline, which estimates scene depth that is suitable for view synthesis tasks. To address the limitations of existing depth estimation methods such as geometric distortions, semantic distortions, and inaccurate depth boundaries, we develop a semantic-aware neural network for depth prediction, couple its estimate with a segmentation-based depth adjustment process, and employ a refinement neural network that facilitates accurate depth predictions at object boundaries. According to this depth estimate, our framework then maps the input image to a point cloud and synthesizes the resulting video frames by rendering the point cloud from the corresponding camera positions. To address disocclusions while maintaining geometrically and temporally coherent synthesis results, we utilize context-aware color- and depth-inpainting to fill in the missing information in the extreme views of the camera path, thus extending the scene geometry of the point cloud. Experiments with a wide variety of image content show that our method enables realistic synthesis results. Our study demonstrates that our system allows users to achieve better results while requiring little effort compared to existing solutions for the 3D Ken Burns effect creation.\vspace{0.5cm}

\end{abstract}



\begin{teaserfigure}\centering
    \setlength{\tabcolsep}{0.05cm}
    \setlength{\itemwidth}{5.89cm}
    \begin{tabular}{ccc}
            \begin{tikzpicture}
                \node [anchor=south west, inner sep=0.0cm] (image) at (0,0) {
                    \includegraphics[width=\itemwidth]{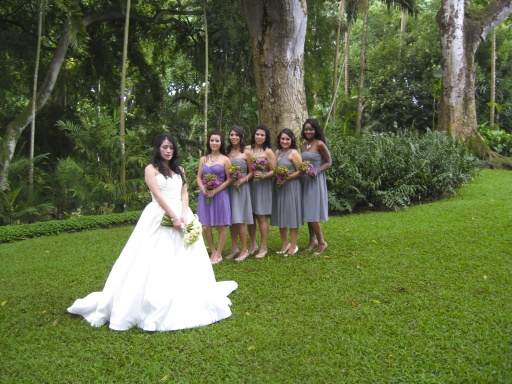}
                };
                \begin{scope}[x={(image.south east)},y={(image.north west)}]
                    \node [anchor=south west, fill=white, inner sep=0.05cm] at (0.02,0.025) {\tiny by Ian D. Keating};
                \end{scope}
            \end{tikzpicture}
        &
            \begin{tikzpicture}
                \definecolor{from}{RGB}{55,135,207}
                \definecolor{to}{RGB}{238,127,14}
                \node [anchor=south west, inner sep=0.0cm] (image) at (0,0) {
                    \includegraphics[width=\itemwidth]{graphics/teaser/1/input}
                };
                \begin{scope}[x={(image.south east)},y={(image.north west)}]
                    \draw [from, line width=0.06cm] (0.03,0.04) rectangle (0.97,0.96);
                    \node [anchor=south west, fill=white] at (0.03 - 0.0105,0.04 - 0.014) {\footnotesize \sc from};
                    \fill [white] (0.97 - 0.0105,0.04 - 0.014) rectangle (0.97 + 0.0105,0.04 + 0.014);
                    \fill [white] (0.03 - 0.0105,0.96 - 0.014) rectangle (0.03 + 0.0105,0.96 + 0.014);
                    \fill [white] (0.97 - 0.0105,0.97 - 0.014) rectangle (0.97 + 0.0105,0.97 + 0.014);
                    \draw [to, line width=0.06cm] (0.12,0.13) rectangle (0.80,0.80);
                    \node [anchor=south west, fill=white] at (0.12 - 0.0105,0.13 - 0.014) {\footnotesize \sc to};
                    \fill [white] (0.80 - 0.0105,0.13 - 0.014) rectangle (0.80 + 0.0105,0.13 + 0.014);
                    \fill [white] (0.12 - 0.0105,0.80 - 0.014) rectangle (0.12 + 0.0105,0.80 + 0.014);
                    \fill [white] (0.80 - 0.0105,0.80 - 0.014) rectangle (0.80 + 0.0105,0.80 + 0.014);
                \end{scope}
            \end{tikzpicture}
        &
            \animategraphics[width=\itemwidth, autoplay, palindrome, final, nomouse, method=widget, poster=last]{15}{graphics/teaser/1/}{00000}{00044}
        \\
            \footnotesize Input Image
        &
            \footnotesize Cropping Windows - User-specified or Automatic
        &
            \footnotesize 3D Ken Burns Video Clip - Using our Tool
        \vspace{0.12cm} \\
            \begin{tikzpicture}
                \node [anchor=south west, inner sep=0.0cm] (image) at (0,0) {
                    \includegraphics[width=\itemwidth]{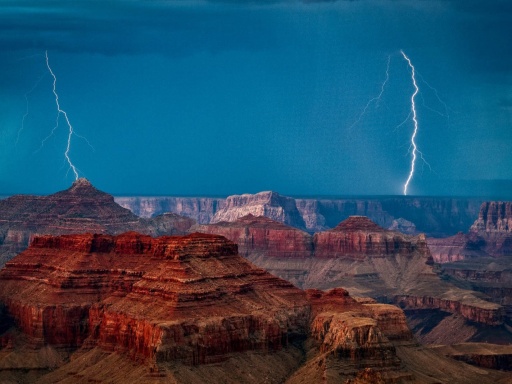}
                };
                \begin{scope}[x={(image.south east)},y={(image.north west)}]
                    \node [anchor=south west, fill=white, inner sep=0.05cm] at (0.02,0.025) {\tiny by Kirk Lougheed};
                \end{scope}
            \end{tikzpicture}
        &
            \begin{tikzpicture}
                \definecolor{from}{RGB}{55,135,207}
                \definecolor{to}{RGB}{238,127,14}
                \node [anchor=south west, inner sep=0.0cm] (image) at (0,0) {
                    \includegraphics[width=\itemwidth]{graphics/teaser/2/input}
                };
                \begin{scope}[x={(image.south east)},y={(image.north west)}]
                    \draw [from, line width=0.06cm] (0.03,0.04) rectangle (0.86,0.86);
                    \node [anchor=south west, fill=white] at (0.03 - 0.0105,0.04 - 0.014) {\footnotesize \sc from};
                    \fill [white] (0.86 - 0.0105,0.04 - 0.014) rectangle (0.86 + 0.0105,0.04 + 0.014);
                    \fill [white] (0.03 - 0.0105,0.86 - 0.014) rectangle (0.03 + 0.0105,0.86 + 0.014);
                    \fill [white] (0.86 - 0.0105,0.86 - 0.014) rectangle (0.86 + 0.0105,0.86 + 0.014);
                    \draw [to, line width=0.06cm] (0.12,0.13) rectangle (0.83,0.82);
                    \node [anchor=south west, fill=white] at (0.12 - 0.0105,0.13 - 0.014) {\footnotesize \sc to};
                    \fill [white] (0.83 - 0.0105,0.13 - 0.014) rectangle (0.83 + 0.0105,0.13 + 0.014);
                    \fill [white] (0.12 - 0.0105,0.82 - 0.014) rectangle (0.12 + 0.0105,0.82 + 0.014);
                    \fill [white] (0.83 - 0.0105,0.82 - 0.014) rectangle (0.83 + 0.0105,0.82 + 0.014);
                \end{scope}
            \end{tikzpicture}
        &
            \animategraphics[width=\itemwidth, autoplay, palindrome, final, nomouse, method=widget, poster=last]{15}{graphics/teaser/2/}{00000}{00044}
        \\
    \end{tabular}\vspace{-0.2cm}
	\caption{3D Ken Burns effect from a single image. Given a single input image and optional user annotations in form of two cropping windows, our framework animates the input image while adding parallax to synthesize a 3D Ken Burns effect. Our method works well for a wide variety of content, including portrait (top) and landscape (bottom) photos. Please refer to our supplementary video demo for more examples. \textbf{Please note that this figure, as well as many other figures in this paper, contain video clips.} Should these videos not already be playing then please consider viewing this paper using Adobe Reader.}\vspace{-0.0cm}
	\label{fig:teaser}
\end{teaserfigure}

\maketitle

\section{Introduction}
\label{sec:introduction}
Advanced image- and video-editing tools allow artists to freely augment photos with depth information and to animate virtual cameras, enabling motion parallax as the camera scans over a still scene. This cinematic effect, which we refer to as 3D Ken Burns effect, has become increasingly popular in documentaries, commercials, and other media. Compared to the traditional Ken Burns effect which animates images with 2D scan and zoom\footnote{\url{http://en.wikipedia.org/wiki/Ken\_Burns\_effect}}, this 3D counterpart enables much more compelling experiences. However, creating such effects from a single image is painstakingly difficult: The photo must be manually separated into different segments, which then have to carefully be arranged in the virtual 3D space, and inpainting needs to be performed to avoid holes when the virtual camera moves away from its origin. In this paper, we target the problem of automatically synthesizing the 3D Ken Burns effect from a single image. We further optionally incorporate simple user-specified camera paths, parameterized by the desired start- and end-view, to grant the user more control over the resulting effect.

This problem of synthesizing realistic moving-camera effects from a single image is highly challenging. Two fundamental concerns need to be addressed. First, to synthesize a new view from a novel camera position, the scene geometry of the original view needs to be recovered accurately. Second, from the predicted scene geometry, a temporally consistent sequence of novel views has to be synthesized which requires dealing with disocclusion. We address both challenges and provide a complete system that enables synthesizing the 3D Ken Burns effect from a single image.

To synthesize the 3D Ken Burns effect, our method first estimates the depth map from the input image. While existing depth prediction methods have rapidly improved over the past few years, monocular depth estimation remains an open problem. We observed that existing depth prediction methods are not particularly suitable for view synthesis tasks such as ours. Specifically, we identified three critical issues of existing depth prediction methods that need to be addressed to make them applicable to 3D Ken Burns synthesis: geometric distortions, semantic distortions, and inaccurate depth boundaries. Based on this observation, we designed a depth estimation pipeline along with the training framework dedicated to addressing these issues. To this end, we developed a semantic-aware neural network for depth estimation and train the network on our newly constructed large-scale synthetic dataset which contains accurate ground truth depth of various photo-realistic scenes.

From the input image and the associated depth map, a sequence of novel views has to be synthesized to produce an output video for the 3D Ken Burns effect. The synthesis process needs to handle three requirements. First, as the camera moves away from its original position, disocclusion necessarily happens. The missing information needs to be filled-in with geometrically consistent content. Second, the novel view renderings need to be synthesized in a temporally consistent manner. The straightforward approach of filling-in the missing information and synthesizing each view independently is not only computationally inefficient but also temporally unstable. Third, we have found that professional artists that use our system manually produce the most compelling effects when they are able to immediately perceive the result of their interaction. The synthesis thus needs to be real-time in order to best support such users. To address these requirements, we propose a simple yet effective solution: We map the input image to points in a point cloud according to the estimated depth. We then perform color- and depth-inpainting of novel view renderings at extreme views like at the beginning and at the end of the virtual camera path. This allows us to extend the point cloud with geometrically sound information. The extended point cloud can then be used to synthesize all novel view renderings in an efficient and temporally consistent manner.

Together, our depth prediction pipeline and novel view synthesis approach provide a complete system for generating the 3D Ken Burns effect from a single image. This system provides a fully automatic solution where the start- and end-view of the virtual camera path are automatically determined so as to minimize the amount of disocclusion. In addition to the fully automatic mode, our system also provides an interactive mode in which users can control the start- and end-view through an intuitive user interface. This allows a more fine-grained control over the resulting 3D Ken Burns effect, thus supporting users in their artistic freedom.

The key contributions of this paper are as follows. We introduce the problem of 3D Ken Burns synthesis from a single image which enables automatic video generation in the form of a moving-camera effect. We leverage existing computer vision technologies and augment them to achieve plausible synthesis results. Our system offers a fully automatic mode which generates a convincing effect without any user feedback, and a view control mode which allows users to control the effect with simple interactions. Experiments on a wide range of real-world imagery demonstrate the effectiveness of our system. Our study shows that our system enables users to achieve better results while requiring little effort compared to existing solutions for the 3D Ken Burns effect creation.

\section{Related Work}
\label{sec:related}
\subsection{Novel View Synthesis}

Novel view synthesis focuses on generating novel views of scenes or 3D objects from input images taken from a sparse set of viewpoints. It is important for a wide range of applications, including virtual and augmented reality~\cite{Hedman_TOG_2017, Huang_OTHER_2017, Rematas_CVPR_2018}, 3D display technologies~\cite{Didyk_TOG_2013, Kellnhofer_TOG_2017, Lai_OTHER_2016, Ranieri_CGF_2012, Xie_ECCV_2016}, and image- or video-manipulation~\cite{Klose_TOG_2015, Kopf_TOG_2016, Lang_TOG_2010, Liu_TOG_2009, Rahaman_TIP_2018, Zitnick_TOG_2004}. Novel view synthesis is typically solved using image based rendering techniques~\cite{Kang_OTHER_2006}, with recent approaches allowing for high-quality view synthesis results~\cite{Chaurasia_TOG_2013, Chaurasia_CGF_2011, Hedman_TOG_2017, Hedone_TOG_2018, Hedtwo_TOG_2018, Penner_TOG_2017}. With the emergence of deep neural networks, learning-based techniques have become an increasingly popular tool for novel view synthesis~\cite{Flynn_CVPR_2016, Ji_CVPR_2017, Kalantari_TOG_2016, Meshry_CVPR_2019, Mildenhall_TOG_2019, Sitzmann_CVPR_2019, Srinivasan_CVPR_2019, Thies_TOG_2019, Thies_ARXIV_2018, Xu_TOG_2019, Zhou_TOG_2018}. To enable high-quality synthesis results, existing methods typically require multiple input views~\cite{Kang_OTHER_2006, Penner_TOG_2017}. In this paper, we target an extreme form of novel view synthesis which aims to generate novel views along the whole camera path given only a single input image.

\subsection{Learning-based View Synthesis from a Single Image}

Recent novel view synthesis methods approach the single-image setting using deep learning~\cite{Tatarchenko_ARXIV_2015, Zhou_ECCV_2016}. Synthesizing novel views from a single image is inherently challenging and existing methods are often only applicable to specific scene types~\cite{Habtegebrial_ARXIV_2018, Liu_CVPR_2018, Nguyen_ARXIV_2019}, 3D object models~\cite{Olszewski_ARXIV_2019, Park_CVPR_2017, Rematas_PAMI_2017, Yan_NIPS_2016, Yang_NIPS_2015}, or domain-specific light field imagery~\cite{Srinivasan_ICCV_2017}. Most relevant to our work are methods that estimate the scene geometry of the input image via depth~\cite{Cun_OTHER_2019, Liu_CVPR_2018}, normal maps~\cite{Liu_CVPR_2018}, or layered depth~\cite{Tulsiani_ECCV_2018}. While we perform depth-based view synthesis as well, we focus on predicting depth maps suitable for high-quality view synthesis. Specifically, we directly improve the estimated depth and thus the estimated scene geometry to suppress artifacts such as geometric distortions and to tailor the depth prediction to the task of view synthesis.

\subsection{Single-image Depth Estimation}

Single-image depth estimation has gained a lot of research interest over the past decades~\cite{Koch_ARXIV_2018}. Recent advances in deep neural networks along with the introduction of annotated depth image datasets~\cite{Abarghouei_CVPR_2018, Chen_NIPS_2016, Laina_OTHER_2016, Li_CVPR_2018, Saxena_PAMI_2009, Silberman_ECCV_2012, Xian_CVPR_2018, Zheng_OTHER_2018} enabled large improvements in monocular depth estimation. Another promising direction is the use of spatial or temporal pixel-correspondence to train for depth estimation in a self-supervised manner~\cite{Garg_ECCV_2016, Godard_CVPR_2017, Gordon_ARXIV_2019, Li_CVPR_2019, Luo_CVPR_2018, Ummenhofer_CVPR_2017, Zhou_CVPR_2017}. However, depth estimation from a single image remains an open research problem. The quality of the predicted depth maps varies depending on the image type and the depth maps from existing methods are in many scenarios not suitable for generating high-quality novel view synthesis results due to geometric and semantic distortions as well as inaccurate depth boundaries. To support the 3D Ken Burns effect synthesis, we develop our depth prediction, adjustment, and refinement to specifically address those issues.

\subsection{Creative Effect Synthesis}

With 3D scene information such as depth or scene layouts, a range of creative camera effects can be produced from the input image, such as depth-of-field synthesis~\cite{Wadhwa_TOG_2018, Wang_TOG_2018}, 2D-to-3D conversion~\cite{Xie_ECCV_2016}, and photo pop-up~\cite{Hoiem_TOG_2005, Srivastava_OTHER_2009}. In this paper, we focus on synthesizing the 3D Ken Burns effect which is a camera motion effect. Our desired output is a whole video corresponding to a given camera path. A number of methods have been proposed in the past to enable camera fly-through effects from a single image. \cite{Horry_OTHER_1997} present a semi-automatic system that lets users represent the scene with a simplified spidery mesh after a manual foreground segmentation process. The image is then projected onto that simplified scene geometry which allows flying a camera through it to obtain certain 3D illusions. Based on a similar idea, follow-up work enriches the scene representation to handle scenes with more than one vanishing point and more diverse camera motions~\cite{Kang_CGF_2001, Li_OTHER_2001}. While realistic effects can be achieved for certain types of images, the simplified scene representation is often too simplistic to handle general types of images and still requires manual segmentation which demands significant user effort. Most related to our work is the system from~\cite{Zheng_OTHER_2009} which synthesizes a video with realistic parallax from still images. This method, however, requires multiple images as input. We focus on a more challenging problem of synthesizing the effect from a single image.

\subsection{Image-to-Video Generation}

The intended output of our method is a video representing the 3D Ken Burns effect. Our research is thus also related to image-to-video generation, an increasingly popular topic in computer vision. Existing work in this area focuses on developing generative models to predict motions in video frames given one or a few starting frames~\cite{Hsieh_NIPS_2018, Lee_ARXIV_2018, Liang_ICCV_2017, Mathieu_ARXIV_2015, Reda_ECCV_2018, Vondrick_NIPS_2016, Xu_CVPR_2018}. While promising results have been achieved for synthesizing object motion in videos with static background, they are often not suitable to synthesize realistic camera motion effects as in our problem.

\section{3D Ken Burns Effect Synthesis}
\label{sec:method}
\begin{figure*}\centering
    \setlength{\tabcolsep}{0.05cm}
    \setlength{\itemwidth}{5.89cm}
    \begin{tabular}{ccc}
            \begin{animateinline}[autoplay, palindrome, final, nomouse, method=widget]{10}
                \multiframe{15}{i=0+1,r=0.0+0.0120}{
                    \begin{tikzpicture}
                        \definecolor{arrowcolor}{RGB}{238,127,14}
                        \node [anchor=south west, inner sep=0.0cm] (image) at (0,0) {
                            \includegraphics[width=\itemwidth]{graphics/depthestim/1-lijun/\zeropad{00000}{\i}}
                        };
                        \begin{scope}[x={(image.south east)},y={(image.north west)}]
                            \node [anchor=south west, fill=white, inner sep=0.05cm] at (0.02,0.025) {\tiny by Leif Skandsen};
                            \draw [double arrow=0.2cm with white and arrowcolor] (0.44 + 0.04 - \r,0.4) -- (0.54 + 0.04 - \r,0.6);
                        \end{scope}
                    \end{tikzpicture}
                }
            \end{animateinline}
        &
            \begin{animateinline}[autoplay, palindrome, final, nomouse, method=widget]{10}
                \multiframe{15}{i=0+1,r=0.0+0.0105}{
                    \begin{tikzpicture}
                        \definecolor{arrowcolor}{RGB}{238,127,14}
                        \node [anchor=south west, inner sep=0.0cm] (image) at (0,0) {
                            \includegraphics[width=\itemwidth]{graphics/depthestim/1-megadepth/\zeropad{00000}{\i}}
                        };
                        \begin{scope}[x={(image.south east)},y={(image.north west)}]
                            \draw [double arrow=0.2cm with white and arrowcolor] (0.44 + 0.03 - \r,0.4) -- (0.54 + 0.035 - \r,0.6);
                        \end{scope}
                    \end{tikzpicture}
                }
            \end{animateinline}
        &
            \begin{animateinline}[autoplay, palindrome, final, nomouse, method=widget]{10}
                \multiframe{15}{i=0+1,r=0.0+0.0055}{
                    \begin{tikzpicture}
                        \definecolor{arrowcolor}{RGB}{97,157,71}
                        \node [anchor=south west, inner sep=0.0cm] (image) at (0,0) {
                            \includegraphics[width=\itemwidth]{graphics/depthestim/1-ours/\zeropad{00000}{\i}}
                        };
                        \begin{scope}[x={(image.south east)},y={(image.north west)}]
                            \draw [double arrow=0.2cm with white and arrowcolor] (0.44 + 0.00 - \r,0.4) -- (0.54 + 0.00 - \r,0.6);
                        \end{scope}
                    \end{tikzpicture}
                }
            \end{animateinline}
        \\
            \footnotesize View synthesis using DeepLens's predicted depth.
        &
            \footnotesize View synthesis using MegaDepth's predicted depth.
        &
            \footnotesize View synthesis using our depth estimation pipeline.
        \vspace{0.12cm} \\
            \begin{animateinline}[autoplay, palindrome, final, nomouse, method=widget]{10}
                \multiframe{15}{i=0+1,r=0.0+0.0160}{
                    \begin{tikzpicture}
                        \definecolor{arrowcolor}{RGB}{238,127,14}
                        \node [anchor=south west, inner sep=0.0cm] (image) at (0,0) {
                            \includegraphics[width=\itemwidth]{graphics/depthestim/2-lijun/\zeropad{00000}{\i}}
                        };
                        \begin{scope}[x={(image.south east)},y={(image.north west)}]
                            \node [anchor=south west, fill=white, inner sep=0.05cm] at (0.02,0.025) {\tiny by Oliver Wang};
                            \draw [double arrow=0.2cm with white and arrowcolor] (0.41 + 0.02 - \r,0.21) -- (0.31 + 0.02 - \r,0.41);
                        \end{scope}
                    \end{tikzpicture}
                }
            \end{animateinline}
        &
            \begin{animateinline}[autoplay, palindrome, final, nomouse, method=widget]{10}
                \multiframe{15}{i=0+1,r=0.0+0.0150}{
                    \begin{tikzpicture}
                        \definecolor{arrowcolor}{RGB}{238,127,14}
                        \node [anchor=south west, inner sep=0.0cm] (image) at (0,0) {
                            \includegraphics[width=\itemwidth]{graphics/depthestim/2-megadepth/\zeropad{00000}{\i}}
                        };
                        \begin{scope}[x={(image.south east)},y={(image.north west)}]
                            \draw [double arrow=0.2cm with white and arrowcolor] (0.41 + 0.00 - \r,0.21) -- (0.31 + 0.00 - \r,0.41);
                            \draw [double arrow=0.2cm with white and arrowcolor] (0.64 + 0.00 - \r,0.91) -- (0.47 + 0.00 - \r,0.81);
                        \end{scope}
                    \end{tikzpicture}
                }
            \end{animateinline}
        &
            \begin{animateinline}[autoplay, palindrome, final, nomouse, method=widget]{10}
                \multiframe{15}{i=0+1,r=0.0+0.0150}{
                    \begin{tikzpicture}
                        \definecolor{arrowcolor}{RGB}{97,157,71}
                        \node [anchor=south west, inner sep=0.0cm] (image) at (0,0) {
                            \includegraphics[width=\itemwidth]{graphics/depthestim/2-ours/\zeropad{00000}{\i}}
                        };
                        \begin{scope}[x={(image.south east)},y={(image.north west)}]
                            \draw [double arrow=0.2cm with white and arrowcolor] (0.41 + 0.00 - \r,0.21) -- (0.31 + 0.00 - \r,0.41);
                            \draw [double arrow=0.2cm with white and arrowcolor] (0.64 + 0.00 - \r,0.91) -- (0.47 + 0.00 - \r,0.81);
                        \end{scope}
                    \end{tikzpicture}
                }
            \end{animateinline}
        \\
    \end{tabular}\vspace{-0.2cm}
    \caption{Geometric- and semantic-distortion examples resulting from off-the-shelf depth estimation methods. These videos were synthesized by moving a virtual camera left and right. To focus the comparison on the depth estimate quality, we do not show our final synthesis result and instead only show the intermediate point-cloud rendering that are subject to disocclusion. In the first row, DeepLens and MegaDepth are subject to geometric distortions in the white building. In the second row, DeepLens and MegaDepth are subject to semantic distortions and are inconsistent with respect to the hand of the boy. Furthermore, MegaDepth's depth prediction also separates the head of the boy from the rest of the body.}\vspace{-0.2cm}
    \label{fig:depthestim}
\end{figure*}

Our framework consists of two main components, namely the depth estimation pipeline (Figure~\ref{fig:overview-depth}), and the novel view synthesis pipeline (Figure~\ref{fig:overview-synthesis}). In this section, we describe each component in detail.

\subsection{Semantic-aware Depth Estimation}

\begin{figure*}\centering
    \setlength{\tabcolsep}{0.05cm}
    \setlength{\itemwidth}{4.4cm}
    \begin{tabular}{cccc}
            \includegraphics[width=\itemwidth]{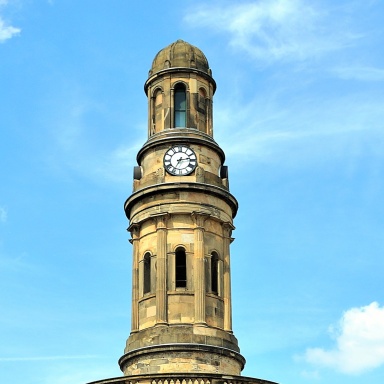}
        &
            \begin{tikzpicture}
                \definecolor{arrowcolor}{RGB}{238,127,14}
                \node [anchor=south west, inner sep=0.0cm] (image) at (0,0) {
                    \includegraphics[width=\itemwidth]{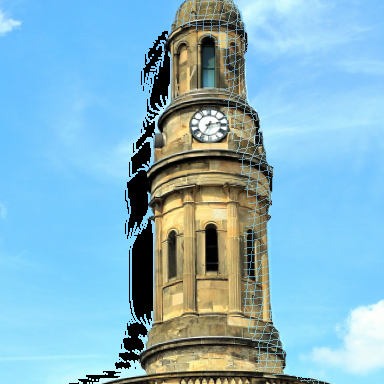}
                };
                \begin{scope}[x={(image.south east)},y={(image.north west)}]
                    \draw [double arrow=0.2cm with white and arrowcolor] (0.1,0.35) -- (0.36,0.40);
                    \draw [double arrow=0.2cm with white and arrowcolor] (0.95,0.7) -- (0.67,0.65);
                \end{scope}
            \end{tikzpicture}
        &
            \begin{tikzpicture}
                \definecolor{arrowcolor}{RGB}{97,157,71}
                \node [anchor=south west, inner sep=0.0cm] (image) at (0,0) {
                    \includegraphics[width=\itemwidth]{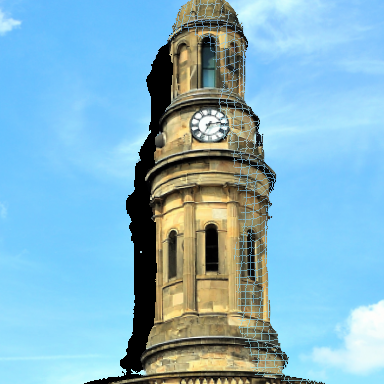}
                };
                \begin{scope}[x={(image.south east)},y={(image.north west)}]
                    \draw [double arrow=0.2cm with white and arrowcolor] (0.1,0.35) -- (0.36,0.40);
                \end{scope}
            \end{tikzpicture}
        &
            \begin{tikzpicture}
                \definecolor{arrowcolor}{RGB}{97,157,71}
                \node [anchor=south west, inner sep=0.0cm] (image) at (0,0) {
                    \includegraphics[width=\itemwidth]{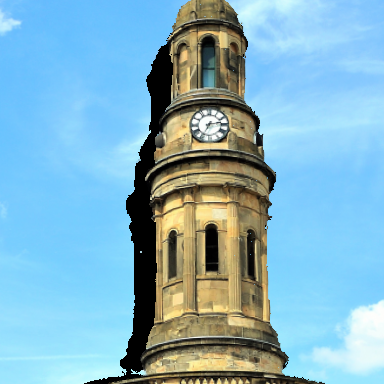}
                };
                \begin{scope}[x={(image.south east)},y={(image.north west)}]
                    \draw [double arrow=0.2cm with white and arrowcolor] (0.95,0.7) -- (0.67,0.65);
                \end{scope}
            \end{tikzpicture}
        \\
            \footnotesize Cropped input image.
        &
            \footnotesize Rendering using initial depth.
        &
            \footnotesize Rendering using refined depth.
        &
            \footnotesize Using refined depth and z-filtering.
        \\
    \end{tabular}\vspace{-0.2cm}
    \caption{Example of our point cloud rendering. Using the point cloud of the initial depth estimate exemplifies the importance of our depth refinement, as objects may otherwise be torn apart at the object boundaries. We further note that moving the virtual camera forward may lead to cracks through which occluded background points may erroneously become visible (note the blue grid pattern on the tower), which we successfully address through z-filtering.}\vspace{-0.2cm}
    \label{fig:rendering}
\end{figure*}
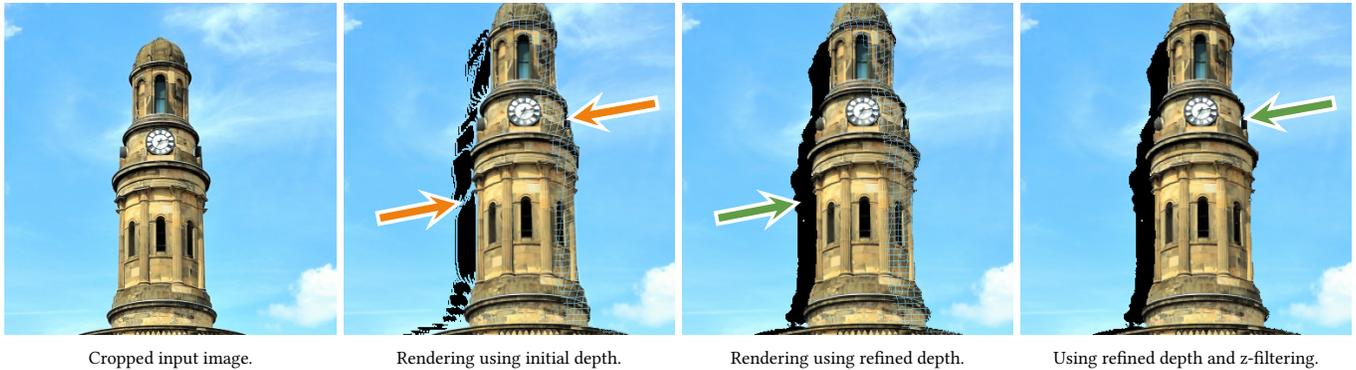

To synthesize the 3D Ken Burns effect, our method first estimates the depth of the input image. While recent advanced methods for monocular depth estimation have shown good performance on public benchmarks, we observed that their predictions are at times not suitable to produce high-quality view synthesis results. In particular, there are at least three major issues when applying existing depth estimation methods to generate the 3D Ken Burns effect:

\begin{enumerate}
    \item \textit{Geometric distortions.} While state-of-the-art depth estimation methods can generate reasonable depth orderings, they often have difficulty in capturing geometric relations such as planarity. Geometric distortion, such as bending planes, thus often appear in the synthesis results (Figure~\ref{fig:depthestim}, top row).
    \item \textit{Semantic distortions.} Existing depth estimation methods predict the depth maps without explicitly taking the semantics of objects into account. Therefore, in many cases the depth values are assigned inconsistently inside regions of the same object, resulting in unnatural synthesis results such as objects sticking to the ground plane or different parts of an object being torn apart (Figure~\ref{fig:depthestim}, bottom row).
    \item \textit{Inaccurate depth boundaries.} Current state-of-the-art methods for single-image depth estimation process the input image at a low resolution and utilize bilinear interpolation to obtain the full-resolution depth estimate. They are thus unable to accurately capture depth boundaries, resulting in artifacts in the novel view renderings (Figure~\ref{fig:rendering}).
\end{enumerate}

\begin{figure*}\centering
    \includegraphics[]{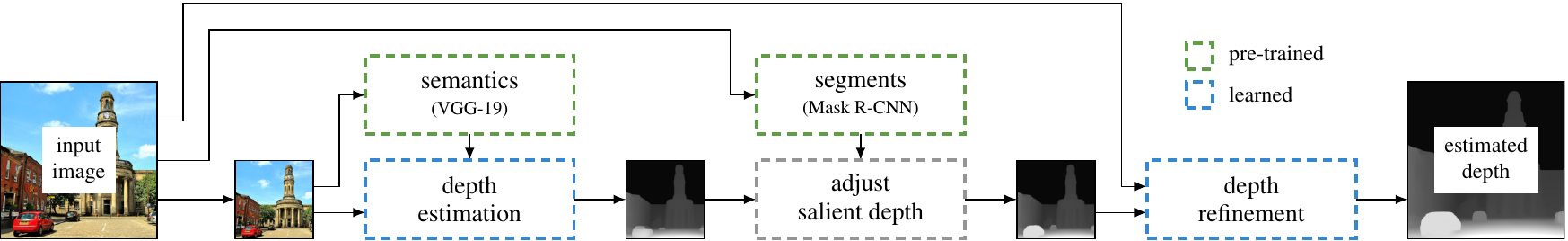}\vspace{-0.0cm}
    \caption{Overview of our depth estimation pipeline. Given a high-resolution image, we start by estimating a coarse depth based on a low-resolution input image. This depth estimation network is guided by semantic information extracted using VGG-19~\cite{Simonyan_ARXIV_2014} and supervised on a computer-generated dataset with accurate ground truth depth in order to facilitate geometrically sound predictions. To avoid semantic distortions, we then adjust the depth map according to the segmentation of Mask R-CNN~\cite{He_ICCV_2017} and make sure that each salient object is mapped to a coherent plane. Lastly, we utilize a depth refinement network that, guided by the input image, upsamples the coarse depth and ensures accurate depth boundaries.}\vspace{-0.1cm}
    \label{fig:overview-depth}
\end{figure*}

In this paper, we design a semantic-aware depth estimation dedicated to addressing these issues. To do so, we separate the depth estimation into three steps. First, estimating coarse depth using a low-resolution image while relying on semantic information extracted using VGG-19~\cite{Simonyan_ARXIV_2014} to facilitate generalizability. Second, adjusting the depth map according to the instance-level segmentation of Mask R-CNN~\cite{He_ICCV_2017} to ensure consistent depth values inside salient objects. Third, refining the depth boundaries guided by the input image while upsampling the low-resolution depth estimate. Our depth estimation pipeline is illustrated in Figure~\ref{fig:overview-depth} and we subsequently elaborate each step.

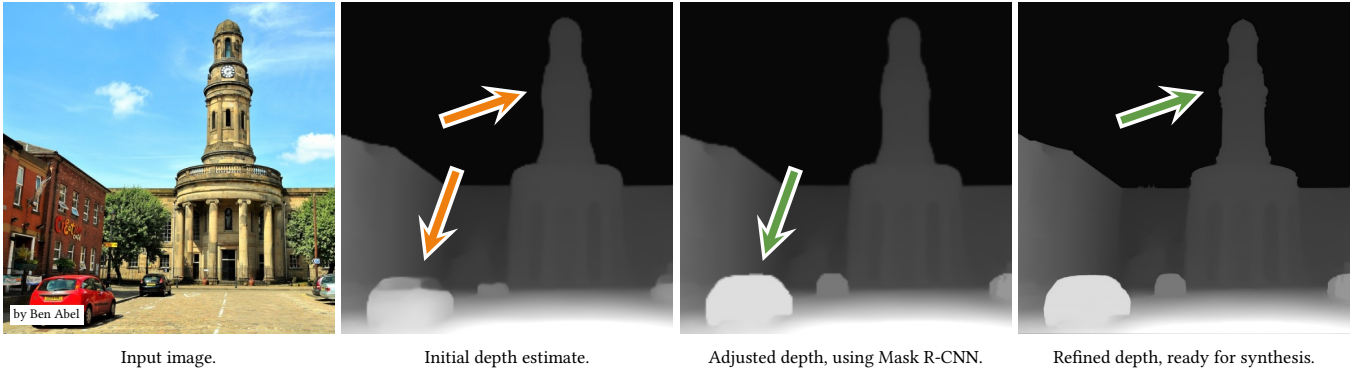
\begin{figure*}\centering
    \setlength{\tabcolsep}{0.05cm}
    \setlength{\itemwidth}{4.4cm}
    \begin{tabular}{cccc}
            \begin{tikzpicture}
                \node [anchor=south west, inner sep=0.0cm] (image) at (0,0) {
                    \includegraphics[width=\itemwidth]{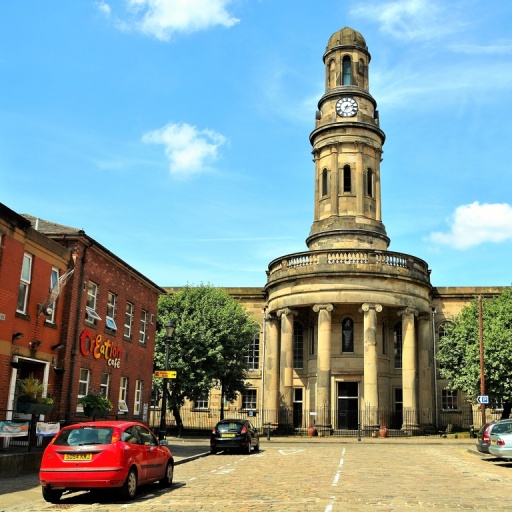}
                };
                \begin{scope}[x={(image.south east)},y={(image.north west)}]
                    \node [anchor=south west, fill=white, inner sep=0.05cm] at (0.02,0.025) {\tiny by Ben Abel};
                \end{scope}
            \end{tikzpicture}
        &
            \begin{tikzpicture}
                \definecolor{arrowcolor}{RGB}{238,127,14}
                \node [anchor=south west, inner sep=0.0cm] (image) at (0,0) {
                    \includegraphics[width=\itemwidth]{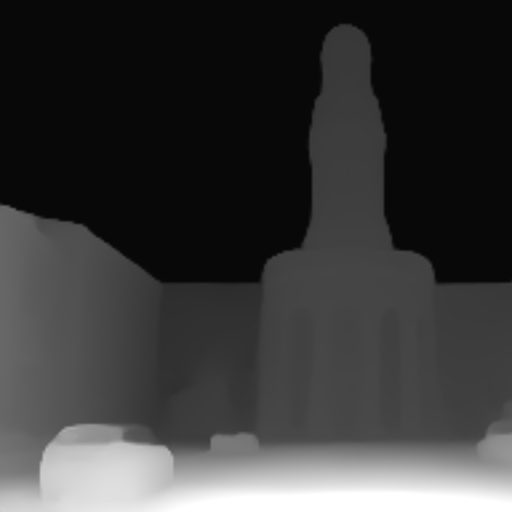}
                };
                \begin{scope}[x={(image.south east)},y={(image.north west)}]
                    \draw [double arrow=0.2cm with white and arrowcolor] (0.35,0.5) --  (0.25,0.22);
                    \draw [double arrow=0.2cm with white and arrowcolor] (0.3,0.64) -- (0.56,0.73);
                \end{scope}
            \end{tikzpicture}
        &
            \begin{tikzpicture}
                \definecolor{arrowcolor}{RGB}{97,157,71}
                \node [anchor=south west, inner sep=0.0cm] (image) at (0,0) {
                    \includegraphics[width=\itemwidth]{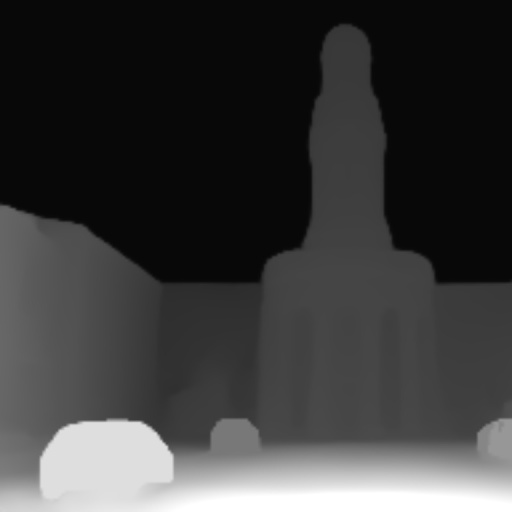}
                };
                \begin{scope}[x={(image.south east)},y={(image.north west)}]
                    \draw [double arrow=0.2cm with white and arrowcolor] (0.35,0.5) --  (0.25,0.22);
                \end{scope}
            \end{tikzpicture}
        &
            \begin{tikzpicture}
                \definecolor{arrowcolor}{RGB}{97,157,71}
                \node [anchor=south west, inner sep=0.0cm] (image) at (0,0) {
                    \includegraphics[width=\itemwidth]{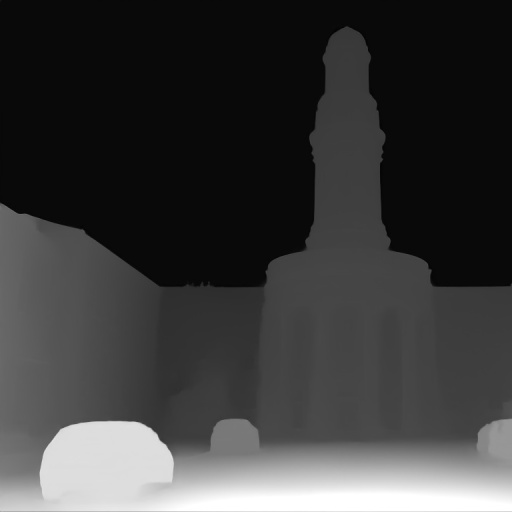}
                };
                \begin{scope}[x={(image.south east)},y={(image.north west)}]
                    \draw [double arrow=0.2cm with white and arrowcolor] (0.3,0.64) -- (0.56,0.73);
                \end{scope}
            \end{tikzpicture}
        \\
            \footnotesize Input image.
        &
            \footnotesize Initial depth estimate.
        &
            \footnotesize Adjusted depth, using Mask R-CNN.
        &
            \footnotesize Refined depth, ready for synthesis.
        \\
    \end{tabular}\vspace{-0.2cm}
    \caption{Intermediate depth estimation results. This example demonstrates the contribution of each stage in our depth estimation pipeline. The initially estimated depth is subject to semantic distortion with respect to the red car and has inaccurate depth boundaries, for example, at the masonry of the tower. The depth adjustment addresses the semantic distortion of the red car, while the depth refinement addresses the fine details at object boundaries.}\vspace{-0.2cm}
    \label{fig:steps}
\end{figure*}

\subsubsection{Depth Estimation}

Following existing work on monocular depth estimation, we leverage a neural network to predict a coarse depth map. To facilitate a semantic-aware depth prediction, we further provide semantic guidance by augmenting the input of our network with the feature maps extracted from the \texttt{pool\_4} layer of VGG-19~\cite{Simonyan_ARXIV_2014}. We found that granting explicit access to this semantic information encourages the network to better capture the geometry of large scene structures, thus addressing the concern of geometric distortions. Different from existing work, we do not resize the input image to a fixed resolution when providing it to the network and instead resize it such that its largest dimension is 512 pixels while preserving its aspect ratio.

\textit{Architecture.} We employ a GridNet~\cite{Fourure_BMVC_2017} architecture with the modifications proposed by \cite{Niklaus_CVPR_2018} to prevent checkerboard artifacts~\cite{Odena_OTHER_2016}. We incorporate this grid architecture with a configuration of six rows and four columns, where the first two columns perform downsampling and the last two columns perform upsampling. This multi-path GridNet architecture allows the network to effectively combine feature representations from multiple scales. We feed the input image into the first row, while inserting the semantic features from VGG-19 into the fourth row of the grid. We explicitly encourage the network to focus more on the semantic features and less on the input image by letting the first three rows of the grid (corresponding to the input image) have a channel size of 32, 48, and 64 respectively while the fourth through sixth row (corresponding to the semantic features) have 512 channels each. As such, a majority of the parameters reside in the bottom half of the network, forcing it to heavily make use of semantic features and in-turn supporting the generalization capability of our depth estimation network. 

\textit{Loss Functions.} To train our depth estimation network, we adopt the pixel-wise $\ell_1$ as well as the scale invariant gradient loss proposed by \cite{Ummenhofer_CVPR_2017} to emphasize depth discontinuities. Specifically, given the ground truth inverse depth $\hat\xi$, we supervise the estimated inverse depth $\xi$ using the $\ell_1$-based loss as
\begin{equation}
    \mathcal{L}_\text{ord} = \textstyle{\sum_{i,j}} \left\| \xi(i,j) - \hat{\xi}(i,j)  \right\| _1
\end{equation}
Similar to~\cite{Ummenhofer_CVPR_2017}, we encourage more pronounced depth discontinuities and stimulate smoothness in homogeneous regions by incorporating a scale invariant gradient loss as
\begin{equation}
    \mathcal{L}_\text{grad} = \sum_{h \in \{1,2,4,8,16\}} \sum_{i,j} \left\| \mathbf{g_h}[\xi](i,j) - \mathbf{g_h}[\hat\xi](i,j) \right\| _2
\end{equation}
where the discrete scale invariant gradient $\mathbf{g}$ is defined as
\begin{equation}
    \mathbf{g_h}[f](i,j) = \left( \tfrac{f(i+h,j) - f(i,j)}{\vert f(i+h,j) \vert + \vert f(i,j) \vert}, \tfrac{f(i,j+h) - f(i,j)}{\vert f(i,j+h) \vert + \vert f(i,j) \vert} \right)^\top
\end{equation}
We emphasize the scale invariant gradient loss when training our depth estimation network and combine the two losses as
\begin{equation}
    \mathcal{L}_\text{depth} = 0.0001 \cdot \mathcal{L}_\text{ord} + \mathcal{L}_\text{grad}
\end{equation}
As such, we encourage accurate depth boundaries which are important when synthesizing the 3D Ken Burns effect.

\textit{Training.} We utilize Adam~\cite{Kingma_ARXIV_2014} with $\alpha = 0.0001$, $\beta_1 = 0.9$, and $\beta_2 = 0.999$ and train our depth estimation network for $3 \cdot 10^6$ iterations. We incorporate 13017 samples from the raw dataset of NYU~v2~\cite{Silberman_ECCV_2012} together with 8685 samples from MegaDepth~\cite{Li_CVPR_2018}. Since these datasets are subject to noise and an inaccurate depth at object boundaries, we also leverage our own dataset which is described in Section~\ref{sec:dataset}. Our dataset consists of realistic renderings which provide high-quality depth maps with clear discontinuities at object boundaries.

\subsubsection{Depth Adjustment}

We have found that our depth prediction network augmented with semantic features and trained using our high-quality dataset significantly improves the scene geometry represented by the estimate depth. However, semantic distortions have not been entirely resolved. It is extremely challenging to obtain accurate object-level depth predictions as the neural network not only needs to reason about the boundary of each object but also needs to determine the geometric relationship between different parts of an object. One approach to address this problem is to either provide semantic labels as input to the depth estimation network, or to train the depth estimation network in a multi-task setting to jointly predict segmentation masks~\cite{Eigen_ICCV_2015, Liu_CVPR_2010, Mousavian_OTHER_2016, Nekrasov_ARXIV_2018} which would encourage the network to reason about object boundaries.

In contrast, we borrow a technique frequently employed by artists when creating the 3D Ken Burns effect manually: Identify the object segments and approximate each object with a frontal plane positioned upright on the ground plane. We mimic this practice and utilize instance-level segmentation masks from Mask R-CNN~\cite{He_ICCV_2017} for this purpose. Specifically, we select the masks of semantically important objects such as humans, cars, and animals and adjust the estimated depth values by assigning the smallest depth value from the bottom of the salient object to the entire mask. We note that this approximation is not physically correct. However, it is effective in producing perceptually plausible results for a majority of content as demonstrated by many artist-created results.

\subsubsection{Depth Refinement}

So far, our depth estimation network is designed to reduce geometric distortions with the depth adjustment addressing semantic distortions. However, the resulting depth estimate is of low resolution and may be erroneous at boundary regions. One possible solution to this problem is to apply joint bilateral filtering to upsample the depth map. However, this does not work well in our case. As also observed in previous work~\cite{Yijunli_ECCV_2016}, we found that the texture of the guiding image tends to be transferred to the upsampled depth. In this work, we thus instead employ a neural network that, guided by a high-resolution image, learns how to perform depth upsampling that is subject to erroneous estimates at object boundaries. During inference, this model predicts the refined depth map at an aspect-dependent resolution with the largest dimension being $1024$ pixels. This upscaling factor can further be increased by modifying the neural network accordingly.

\textit{Architecture.} We insert the input image into a U-Net with three downsampling blocks which use strided convolutions and three corresponding upsampling blocks which use convolutions and bilinear upsampling. We insert the estimated depth at the bottom of the U-Net, allowing the network to learn how to downsample the input image in order to guide the depth during upsampling.

\textit{Loss Functions.} Like with our depth estimation network, we encourage accurate predictions at object boundaries and employ the same $\mathcal{L}_\text{depth}$ loss when training our refinement network.

\textit{Training.} We utilize Adam~\cite{Kingma_ARXIV_2014} with $\alpha = 0.0001$, $\beta_1 = 0.9$, and $\beta_2 = 0.999$ and train our depth refinement network for $1 \cdot 10^6$ iterations. Since accurate ground truth depth boundaries are crucial for training this network, we only use our computer-generated dataset which is described in Section~\ref{sec:dataset}. Specifically, we downsample and distort the ground truth depth to simulate the coarse predicted depth map and use it, together with the high-resolution image, as inputs to the depth refinement network.

\subsubsection{Summary}

Our depth estimation pipeline is designed to address each of the identified issues that are important when using depth estimation methods to create the 3D Ken Burns effect: geometric distortions, semantic distortions, and inaccurate depth boundaries. Please see Figure~\ref{fig:steps} which demonstrates the contribution of each step in our pipeline to the final depth estimate.

\begin{figure*}\centering
    \setlength{\tabcolsep}{0.05cm}
    \setlength{\itemwidth}{4.4cm}
    \begin{tabular}{cccc}
            \animategraphics[width=\itemwidth, autoplay, palindrome, final, nomouse, method=widget]{10}{graphics/inpainting/1-none/}{00000}{00014}
        &
            \animategraphics[width=\itemwidth, autoplay, palindrome, final, nomouse, method=widget]{10}{graphics/inpainting/1-deepfill/}{00000}{00014}
        &
            \animategraphics[width=\itemwidth, autoplay, palindrome, final, nomouse, method=widget]{10}{graphics/inpainting/1-edgeconnect/}{00000}{00014}
        &
            \animategraphics[width=\itemwidth, autoplay, palindrome, final, nomouse, method=widget]{10}{graphics/inpainting/1-ours/}{00000}{00014}
        \\
            \footnotesize Without inpainting.
        &
            \footnotesize Using DeepFill~\cite{Yu_CVPR_2018}.
        &
            \footnotesize Using EdgeConnect~\cite{Nazeri_ARXIV_2019}.
        &
            \footnotesize Our inpainting.
        \\
    \end{tabular}\vspace{-0.2cm}
    \caption{Example video synthesis results, comparing two popular off-the-shelf inpainting methods with our approach. DeepFill fails to inpaint a plausible result due to the non-rectangular nature of the area that is ought to be inpainted. EdgeConnect inpaints a more plausible result but is not temporally consistent and fails to preserve the object boundary. In contrast, our inpainting approach is both temporally consistent and maintains a clear object boundary.}\vspace{-0.1cm}
    \label{fig:inpainting}
\end{figure*}

\begin{figure*}\centering
    \includegraphics[]{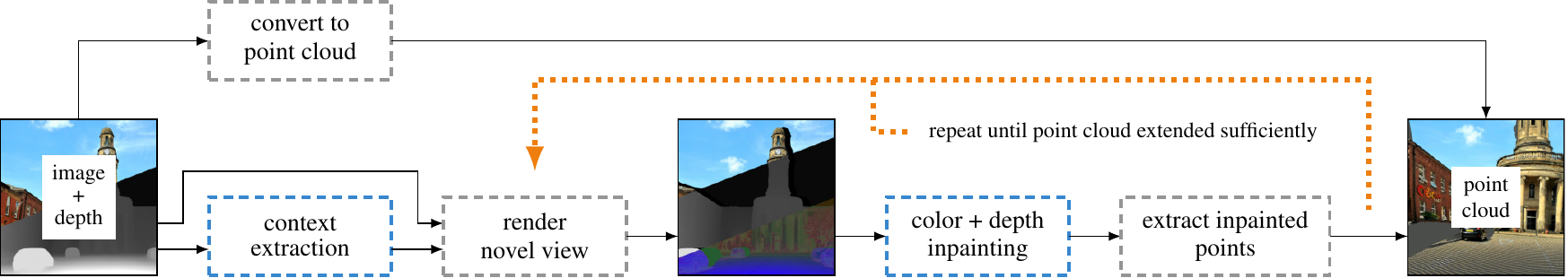}\vspace{-0.0cm}
    \caption{Overview of our novel view synthesis approach. From the point cloud obtained from the input image and the estimated depth map, we render consecutive novel views from new camera positions. This point cloud is only a partial view of the world geometry though, which is why novel view renderings will be subject to disocclusion. To address this issue, we perform geometrically consistent color- and depth-inpainting to recover a complete novel view from an incomplete render where each pixel contains color-, depth-, and context-information. The inpainted depth can then be used to map the inpainted color to new points in the existing point cloud. By repeating this procedure until the point cloud has been extended sufficiently, it is possible to render complete and temporally consistent novel views in real time. To synthesize the 3D Ken Burns effect along a camera path, it is in this regard sufficient to perform the color- and depth-inpainting only at extreme views like at the beginning and at the end.}\vspace{-0.2cm}
    \label{fig:overview-synthesis}
\end{figure*}

\subsection{Context-aware Inpainting for View Synthesis}

To synthesize the 3D Ken Burns effect from the estimated depth, our method first maps the input image to points in a point cloud. Each frame of the resulting video can then be synthesized by rendering the point cloud from the corresponding camera position along a pre-determined camera path. The point cloud, however, is only a partial view of the world geometry as seen from the input image. Therefore, the resulting novel view renderings are incomplete with holes caused by disocclusion. One possible solution is to utilize off-the-shelf image inpainting methods to fill-in the missing areas in each synthesized video frame. This approach, however, fails to satisfy the following requirements:

\begin{enumerate}
    \item \textit{Geometrically consistent inpainting.} Due to the nature of disocclusion, the filled-in area should resemble the background with a clear separation of the foreground object. Existing off-the-shelf inpainting methods do not explicitly reason about the geometry of the inpainting result though, which is why they are unable to satisfy this requirement (Figure~\ref{fig:inpainting}).
    \item \textit{Temporal consistency.} When rendering multiple novel views to generate a moving-camera effect, the result needs to be temporally consistent. The traditional inpainting formulation does not consider our given scenario, which is why independently applying an existing off-the-shelf inpainting method is subject to temporal inconsistencies (Figure~\ref{fig:inpainting}).
    \item \textit{Real-time synthesis.} When manually specifying the camera path for the 3D Ken Burns effect, we found that the best user experience is achieved when users can immediately perceive the result and make adjustments accordingly. Applying off-the-shelf inpainting methods in a frame-by-frame manner would be too computationally expensive to adequately support this use case scenario (Section~\ref{sec:interface}).
\end{enumerate}

In this paper, we design a dedicated view synthesis pipeline to address these requirements as illustrated in Figure~\ref{fig:overview-synthesis}. Given the point cloud obtained from the input image and its depth estimate, we perform joint color- and depth-inpainting to fill-in missing areas in incomplete novel view renderings. Having the inpainting method also incorporate depth enables geometrically consistent inpainting. The inpainted depth can then be used to map the inpainted color to new points in the existing point cloud, addressing the problem of disocclusion. To synthesize the 3D Ken Burns effect along a pre-determined camera path, it is in this regard sufficient to perform the color- and depth-inpainting only at extreme views like at the beginning and at the end. Rendering this extended point cloud preserves temporal consistency and can be done in real-time. To enable real-time synthesis when having an artist specify an arbitrary camera path, we repeat this procedure at extreme views to the left, right, top, and bottom. Our synthesis approach is illustrated in Figure~\ref{fig:overview-synthesis} and we subsequently elaborate the involved steps.

\subsubsection{Point Cloud Rendering} We obtain novel view renderings by projecting the point cloud to an image plane subject to the pinhole camera model. In doing so, we utilize a z-buffer to correctly address occlusion. When moving the virtual camera forward, the point cloud rendering may, however, suffer from shine-through artifacts in which occluded background points becomes visible in foreground regions. \cite{Tulsiani_ECCV_2018} address these artifacts by rendering the point cloud at half the input resolution. In order to preserve the image resolution, we instead filter the z-buffer before projecting the points to the image plane. Specifically, we identify shined-through artifact regions by identifying pixels for which two adjacently opposing neighbors are significantly closer to the virtual camera. We then fill the cracks in the z-buffer with the average depth of the neighboring foreground pixels.

\subsubsection{Context Extraction}

\cite{Niklaus_CVPR_2018} observed that incorporating contextual information is beneficial for generating high-quality novel view synthesis results. Specifically, each point in the point cloud can be extended with contextual information that describes the neighborhood of where the corresponding pixel used to be in the input image. This augments the point cloud with rich information that can, for example, be leveraged for computer graphics in the form of neural rendering~\cite{Aliev_ARXIV_2019, Bui_OTHER_2018, Meshry_CVPR_2019}. To make use of this technique, we leverage a neural network with two convolutional layers to extract 64 channels of context information from the input image. We train this context extractor jointly with the subsequent inpainting network, which allows the extractor to learn how to gather information that is useful when inpainting incomplete novel view renderings.

\subsubsection{Color- and Depth-inpainting}

Different from existing image inpainting methods, our inpainting network accepts color-, depth-, and context-information as input and performs joint color- and depth-inpainting. The additional context provides rich information that is beneficial for high-quality image synthesis while the depth enables geometrically consistent inpainting results with foreground objects clearly being separated from the background. Specifically, we render the color-, depth-, and context-information of the input image to a novel view that is incomplete due to disocclusion. We then use our color- and depth-inpainting network to fill-in missing areas. The inpainted depth allows us to map the inpainted color to new points in the existing point cloud, effectively extending the world geometry that the point cloud represents.

\begin{figure}\centering
    \setlength{\tabcolsep}{0.05cm}
    \setlength{\itemwidth}{4.19cm}
    \begin{tabular}{cc}
            \includegraphics[width=\itemwidth]{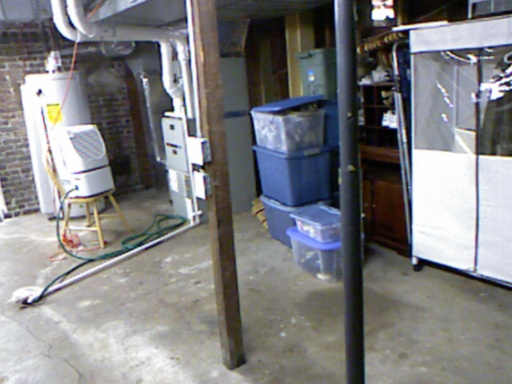}
        &
            \includegraphics[width=\itemwidth]{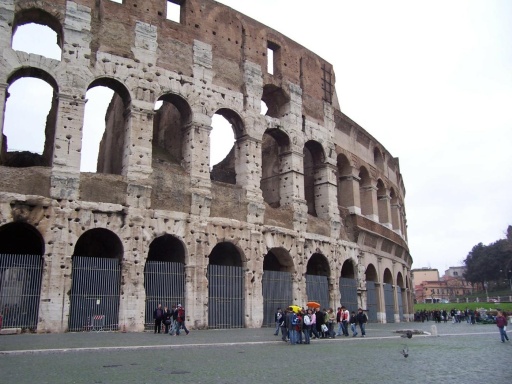}
        \\
            \includegraphics[width=\itemwidth]{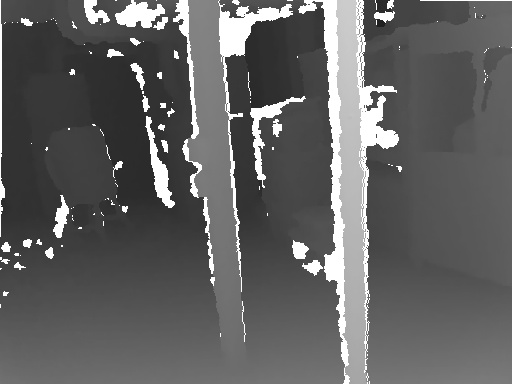}
        &
            \includegraphics[width=\itemwidth]{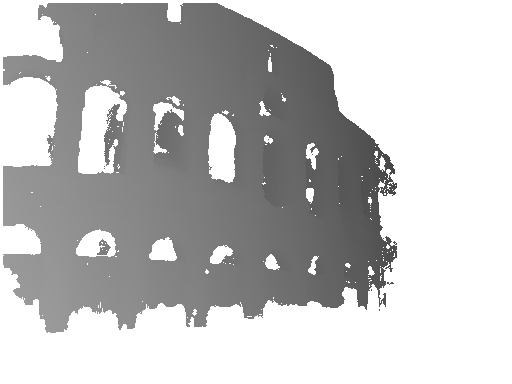}
        \\
            \footnotesize Sample from the NYU~v2 depth dataset.
        &
            \footnotesize Sample from the MegaDepth dataset.
        \\
    \end{tabular}\vspace{-0.2cm}
    \caption{Examples from the NYU~v2 and the MegaDepth dataset, which provide sparse annotations that are subject to inaccurate depth boundaries.}\vspace{-0.2cm}
    \label{fig:otherdata}
\end{figure}

\textit{Architecture.} Similarly to our depth estimation network, we employ a GridNet~\cite{Fourure_BMVC_2017} architecture for our inpainting network due to its ability to learn how to combine representations at multiple scales. Specifically, we utilize a grid with four rows and four columns with a per-row channel size of 32, 64, 128, and 256 respectively. It accepts the color, depth, and context of the incomplete novel view rendering and returns the inpainted color and depth.

\textit{Loss Functions.} We adopt a pixel-wise $\ell_1$ loss as well as a perceptual loss based on deep image features to supervise the color inpainting. Specifically, given a ground truth novel view $I_{gt}$, we supervise the inpainted color $I$ using the $\ell_1$-based loss as
\begin{equation}
    \mathcal{L}_\text{color} = \left\| I - I_{gt} \right\| _1^{\vphantom{1}}
\end{equation}
For the perceptual loss, we employ a content loss based on the difference between deep image features as
\begin{equation}
    \mathcal{L}_\text{percep} = \left\| \phi(I) - \phi(I_{gt}) \right\| _2^2
\end{equation}
where $\phi$ represents feature activations from a generic image classification network. Specifically, we use the activations of the \verb|relu4_4| layer from VGG-19~\cite{Simonyan_ARXIV_2014}. To supervise the depth-inpainting, we use the $\ell_1$-based loss $\mathcal{L}_\text{ord}$ as well as the scale invariant gradient loss $\mathcal{L}_\text{grad}$, thus yielding
\begin{equation}
    \mathcal{L}_\text{inpaint} = \mathcal{L}_\text{color} + \mathcal{L}_\text{percep} + 0.0001 \cdot \mathcal{L}_\text{ord} + \mathcal{L}_\text{grad}
\end{equation}
as the combination of loss functions that we use to supervise the training of our color- and depth-inpainting network.

\textit{Training.} We utilize Adam~\cite{Kingma_ARXIV_2014} with $\alpha = 0.0001$, $\beta_1 = 0.9$ and $\beta_2 = 0.999$ and train our inpainting network for $2 \cdot 10^6$ iterations. Given an input image, we require ground truth novel views to supervise the training of the inpainting network. To this end, we extended our synthetic dataset and collected multiple views as described in Section~\ref{sec:dataset} and shown in Figure~\ref{fig:dataset}.

\subsubsection{Summary} Our novel view synthesis approach is designed to address each of the identified requirements that are important when synthesizing the 3D Ken Burns effect: geometrically consistent inpainting, temporal consistency, and real-time synthesis. Please consider our supplementary video demo to further examine our synthesis results. This video demo also contains an example interaction with our user interface which exemplifies why real-time synthesis is a key feature when manually specifying the camera path.

\begin{figure}\centering
    \setlength{\tabcolsep}{0.05cm}
    \setlength{\itemwidth}{4.19cm}
    \begin{tabular}{cc}
            \animategraphics[width=\itemwidth, autoplay, loop, final, nomouse, method=widget]{1}{graphics/dataset-helium/rgb/}{00000}{00003}
        &
            \animategraphics[width=\itemwidth, autoplay, loop, final, nomouse, method=widget]{1}{graphics/dataset-helium/depth/}{00000}{00003}
        \\
            \footnotesize Sample from our training dataset.
        &
            \footnotesize Corresponding ground truth depth.
        \\
    \end{tabular}\vspace{-0.2cm}
    \caption{Example sequence of four neighboring views from our training dataset. It is computer generated and consists of 134041 scene captures with 4 views each from 32 photo-realistic environments.}\vspace{-0.2cm}
    \label{fig:dataset}
\end{figure}

\subsection{User Interface}
\label{sec:interface}

Given an input image, our system synthesizes the 3D Ken Burns effect from a virtual camera path parameterized by a start- and end-position. We obtain a sequence of frames by uniformly sampling novel view renderings across the linear path between the two positions. Here we describe how to derive camera positions from cropping windows placed on the input image, how to automatically select suitable cropping windows, and how to support the artist in using our system interactively.

\subsubsection{Camera Parametrization} When synthesizing the 2D Ken Burns effect, it is common practice to specify a source- and a target-crop within the input image. This approach provides an intuitive way to manually define the 2D scan and zoom. We adopt this paradigm of parameterizing the start- and end-view for our 3D Ken Burns effect. It is not trivial to match a cropping window in the 2D image space to a virtual camera position in 3D space. In our method, we choose the XY-coordinate of the two virtual cameras such that the foreground object within the scene moves in accordance with the cropping windows. That is, if the source- and target-crop are 100 pixels apart then the foreground object should move by 100 pixels in the synthesized 3D Ken Burns result. Lastly, we use the size of the cropping windows in relation to the input image to determine the Z-coordinate of the corresponding virtual cameras.

\subsubsection{Automatic Mode}
\label{sec:automode}
In the fully automatic mode, we let the algorithm automatically determine the start- and end-view such that the amount of disocclusion is minimized. Specifically, we treat the entire input image as the start-view and employ a uniform sampling grid to find the cropping window corresponding to the end-view that results in the minimum amount of disocclusion. In the resulting 3D Ken Burns effect, the virtual camera naturally approaches the the dominant salient foreground object and emphasizes it through motion parallax. An example result that we obtained using the automatic mode can be found at the top of Figure~\ref{fig:teaser}.

\subsubsection{Interactive Mode} Some users may desire a more fine-grained control over the synthesized 3D Ken Burns effect. To support this use case, we provide an interactive mode in which users determine the two cropping windows which represent the start- and end-view. Thanks to our efficient novel view rendering pipeline, our system can provide real-time feedback when manipulating the start- and end-view windows, which allows users to immediately perceive the effect of their actions. Please refer to our supplementary video demo for an example of our system in action.

\subsection{Training Data}
\label{sec:dataset}

We evaluated several datasets that provide ground truth depth information to supervise the training of our depth estimation pipeline, including the MegaDepth~\cite{Li_CVPR_2018} as well as the NYU~v2~\cite{Silberman_ECCV_2012} dataset. However, as shown in Figure~\ref{fig:otherdata}, these datasets only provide sparse annotations that are subject to inaccurate depth boundaries. We also examined the KITTI dataset~\cite{Geiger_OTHER_2013}, which also provides multi-view data and thus would be useful to supervise the training of our color- and depth-inpainting network. However, it is sparse and subject to inaccuracies as well and particularly limited in terms of scene types and content. As previously shown in Figure~\ref{fig:rendering}, accurate depth boundaries are crucial for novel view synthesis.

We thus created our own computer-generated dataset from 32 virtual environments, which enables us to extract accurate ground truth depth information. Those virtual environments were collected from the UE4 Marketplace\footnote{\url{http://www.unrealengine.com/marketplace/en-US/store}}. We intentionally collected highly realistic environments covering a wide range of scene types such as indoor scenes, urban scenes, rural scenes, and nature scenes. More specifically, we use the Unreal Engine to create a virtual camera rig to capture 134041 scenes from 32 environments where each scene consists of 4 views. Each view contains color-, depth-, and normal-maps at a resolution of $512 \times 512$ pixels. Please see Figure~\ref{fig:dataset} for an example from our dataset. While we did not use any normal-maps, we collected them regardless such that other researchers can make better use of our dataset in the future. Note that, while training our depth estimation network, we randomly crop either the top and bottom or the left and right of each sample in order to facilitate invariance to the aspect ratio of the input image.

\section{Experiments}
\label{sec:experiements}
\begin{figure}\centering
    \hspace{-0.2cm}\includegraphics[]{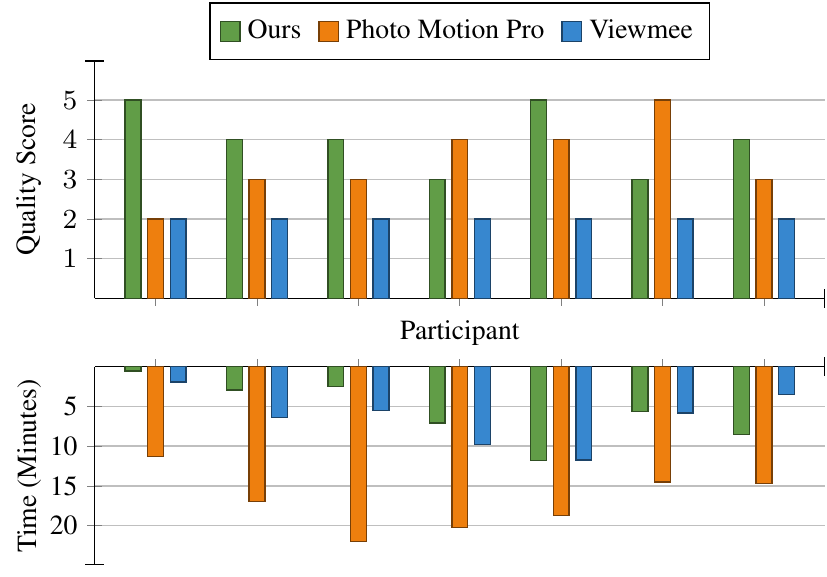}\vspace{-0.2cm}
	\caption{Usability study results. Our study shows that our system enables users to achieve good results while requiring much less effort.}\vspace{-0.2cm}
	\label{fig:user-study}
\end{figure}

\subsection{Usability Study}
\label{sec:usability}

We conduct an informal user study to evaluate the usability of our system in supporting the creation of the 3D Ken Burns effect. In particular, we are interested in investigating how easy it is for non-expert users to achieve desirable results for images with different content. To simulate a plausible scenario, we collected 3D Ken Burns videos created by artists. Specifically, we searched for phrases like ``3D Ken Burns effect'' or ``Parallax Effect'' on YouTube and selected 30 representative results from tutorial videos. We then only further considered those results that do not contain additional artistic effects such as compositing, artificial lighting, and particle effects. We categorize the remaining videos into four groups according to the scene types of the input image, namely ``landscape'', ``portrait'', ``indoor'', ``man-made outdoor environment'' and randomly selected three videos in each category. We thus conduct our informal user study on those 12 examples, for which we have the input image as well as reference 3D Ken Burns effect results.

\begin{figure*}\centering
    \setlength{\tabcolsep}{0.05cm}
    \setlength{\itemwidth}{5.89cm}
    \begin{tabular}{ccc}
            \includegraphics[width=\itemwidth, trim={0.0cm 0.0cm 0.0cm 0.7cm}, clip]{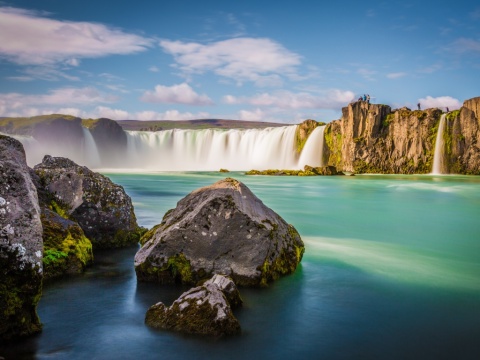}
        &
            \animategraphics[width=\itemwidth, trim={0.0cm 0.0cm 0.0cm 0.7cm}, autoplay, palindrome, final, nomouse, method=widget, poster=last]{10}{graphics/twodee/AdobeStock_69636453-two/}{00000}{00014}
        &
            \animategraphics[width=\itemwidth, trim={0.0cm 0.0cm 0.0cm 0.7cm}, autoplay, palindrome, final, nomouse, method=widget, poster=last]{10}{graphics/twodee/AdobeStock_69636453-ours/}{00000}{00014}
        \\
            \footnotesize Input image.
        &
            \footnotesize 2D Ken Burns with scan and zoom.
        &
            \footnotesize 3D Ken Burns from our system.
        \vspace{0.12cm} \\
            \includegraphics[width=\itemwidth, trim={0.0cm 0.0cm 0.0cm 0.7cm}, clip]{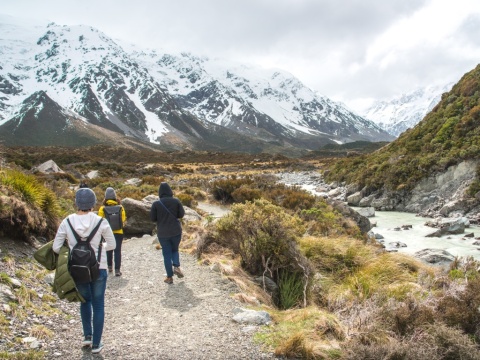}
        &
            \animategraphics[width=\itemwidth, trim={0.0cm 0.0cm 0.0cm 0.7cm}, autoplay, palindrome, final, nomouse, method=widget, poster=last]{10}{graphics/twodee/AdobeStock_219382517-two/}{00000}{00014}
        &
            \animategraphics[width=\itemwidth, trim={0.0cm 0.0cm 0.0cm 0.7cm}, autoplay, palindrome, final, nomouse, method=widget, poster=last]{10}{graphics/twodee/AdobeStock_219382517-ours/}{00000}{00014}
        \\
    \end{tabular}\vspace{-0.2cm}
	\caption{Example results comparing the common 2D Ken Burns with our 3D Ken Burns approach. Notice the difference in motion parallax.}\vspace{-0.2cm}
	\label{fig:twodee}
\end{figure*}

\begin{figure}\centering
    \includegraphics[]{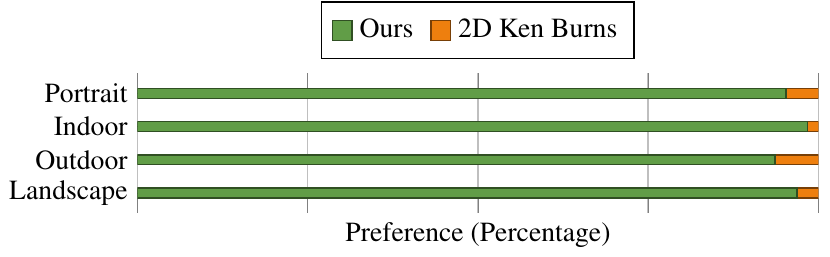}\vspace{-0.2cm}
	\caption{Results from a subjective user study comparing our 3D Ken Burns synthesis to a 2D baseline, indicating a strong preference for our system.}\vspace{-0.2cm}
	\label{fig:user-twodee}
\end{figure}

We recruit 8 participants for our study. In each session, the participant is assigned one image along with the reference result created by an artist. The participant is asked to use our as well as two other systems to create a similar effect from the provided image. The order in which the systems are being used is randomly selected for each participant. The usability and quality of each tool is subjectively rated by the participant at the end of the session.

We compare our framework with existing solutions for creating the 3D Ken Burns effect. We consider two commercial systems. The first is the Photo Motion software package\footnote{\url{http://www.videohive.net/item/photo-motion-pro/13922688}} which is implemented as a template for Adobe After Effects\footnote{\url{http://www.adobe.com/products/aftereffects.html}}. This package provides a commercial implementation for the framework introduced by~\cite{Horry_OTHER_1997} which is one of the most well-known frameworks for interactive camera fly-through synthesis. The second baseline system we consider is the mobile app Viewmee\footnote{\url{http://itunes.apple.com/us/app/id1222280873}} that has been developed to allow non-expert users to easily create the 3D Ken Burns effect. This is one of very few systems that support simple interactions targeting casual users with limited image- or video-editing experience. 

At the end of each session, the participant is asked to rate the three systems in terms of two criteria: system usability and result quality. For system usability, the participant rates each system with a score from one to five, with one indicating the lowest usability (i.e. the tool is too difficult to use to obtain acceptable results within the allocated 30 minutes) and five indicating the best usability (i.e. the tool is easy to use to create good results). For the result quality, the participant is shown the three results that he or she created and asked to score each result from one to five, with one indicating the lowest quality and five indicating the highest quality.

We compare the user-provided usability scores as well as the per-system time for each of the 8 participants in Figure~\ref{fig:user-study}. The results show that using our system, the participants can obtain better results with much less effort compared to the other systems. Viewmee only seems to work for cases with a distinct foreground object in front of a distant background. Photo Motion Pro can model the scene depth for scenes with clear perspective but requires a lot of effort for manual segmentation and scene arrangement. It also is extremely difficult to use in scenes with many different depth layers. Please refer to our supplementary materials for more visual examples shown in video form.

\begin{figure*}\centering
	\setlength{\tabcolsep}{0.0cm}
	\renewcommand{\arraystretch}{1.35}
	\newcommand{\quantTit}[2]{\multicolumn{#1}{c}{\scriptsize #2}}
	\newcommand{\quantSec}[1]{\scriptsize #1}
	\newcommand{\quantInd}[1]{\scriptsize #1}
	\newcommand{\quantVal}[1]{\scalebox{0.9}[1.0]{$ #1 $}}
	\newcommand{\quantBes}[1]{\scalebox{0.9}[1.0]{$\uline{ #1 }$}}
	\footnotesize
	\begin{tabularx}{\textwidth}{@{\hspace{0.1cm}} X p{2.3cm} P{1.0cm} @{\hspace{-0.4cm}} P{1.0cm} @{\hspace{-0.4cm}} P{1.0cm} @{\hspace{-0.4cm}} P{1.0cm} @{\hspace{-0.4cm}} P{1.0cm} @{\hspace{-0.4cm}} P{1.0cm} P{1.0cm} @{\hspace{-0.4cm}} P{1.0cm} @{\hspace{-0.4cm}} P{1.0cm} @{\hspace{-0.4cm}} P{1.0cm} @{\hspace{-0.4cm}} P{1.0cm} @{\hspace{-0.4cm}} P{1.0cm} @{\hspace{-0.07cm}} P{1.0cm} @{\hspace{-0.4cm}} P{1.0cm} @{\hspace{-0.07cm}} P{1.0cm} @{\hspace{-0.4cm}} P{1.0cm} @{\hspace{-0.07cm}} P{1.0cm} @{\hspace{-0.4cm}} P{1.0cm} @{\hspace{-0.4cm}} P{1.0cm} @{\hspace{-0.07cm}}}
		\toprule
			& & \multicolumn{6}{c}{\begin{tikzpicture}\draw [] (0,0.125) -- (0,-0.125); \draw [] (3.8,0.125) -- (3.8,-0.125); \draw [densely dashed] (0,0) -- (3.8,0); \node [fill=white] at (1.9,0) {\normalsize\texttt{NYU-v2}};\end{tikzpicture}} & \multicolumn{13}{c}{\begin{tikzpicture}\draw [] (0,0.125) -- (0,-0.125); \draw [] (8.8,0.125) -- (8.8,-0.125); \draw [densely dashed] (0,0) -- (8.8,0); \node [fill=white] at (4.4,0) {\normalsize\texttt{iBims-1}};\end{tikzpicture}}
		\\
			& & \quantTit{6}{Standard Metrics ($\sigma_i = 1.25^i$)} & \quantTit{6}{Standard Metrics ($\sigma_i = 1.25^i$)} & \quantTit{2}{PE (cm / deg)} & \quantTit{2}{DBE (px)} & \quantTit{3}{DDE (\% for $d = 3$ m)}
		\\ \cmidrule(l{2pt}r{2pt}){3-8} \cmidrule(l{2pt}r{2pt}){9-14} \cmidrule(l{2pt}r{2pt}){15-16} \cmidrule(l{2pt}r{2pt}){17-18} \cmidrule(l{2pt}r{2pt}){19-21}
			& & \quantSec{rel} & \quantSec{log10} & \quantSec{RMS} & \quantSec{$\sigma_1$} & \quantSec{$\sigma_2$} & \quantSec{$\sigma_3$} & \quantSec{rel} & \quantSec{log10} & \quantSec{RMS} & \quantSec{$\sigma_1$} & \quantSec{$\sigma_2$} & \quantSec{$\sigma_3$} & \quantSec{$\varepsilon^\text{plan}_\text{PE}$} & \quantSec{$\varepsilon^\text{orie}_\text{PE}$} & \quantSec{$\varepsilon^\text{acc}_\text{DBE}$} & \quantSec{$\varepsilon^\text{comp}_\text{DBE}$} & \quantSec{$\varepsilon^{0}_\text{DDE}$} & \quantSec{$\varepsilon^{+}_\text{DDE}$} & \quantSec{$\varepsilon^{-}_\text{DDE}$}
		\\
			Method & Training Data & \quantInd{$\downarrow$} & \quantInd{$\downarrow$} & \quantInd{$\downarrow$} & \quantInd{$\uparrow$} & \quantInd{$\uparrow$} & \quantInd{$\uparrow$} & \quantInd{$\downarrow$} & \quantInd{$\downarrow$} & \quantInd{$\downarrow$} & \quantInd{$\uparrow$} & \quantInd{$\uparrow$} & \quantInd{$\uparrow$} & \quantInd{$\downarrow$} & \quantInd{$\downarrow$} & \quantInd{$\downarrow$} & \quantInd{$\downarrow$} & \quantInd{$\uparrow$} & \quantInd{$\downarrow$} & \quantInd{$\downarrow$}
		\\ \cmidrule(l{2pt}r{6pt}){1-1} \cmidrule(l{-2pt}r{2pt}){2-2} \cmidrule(l{2pt}r{2pt}){3-8} \cmidrule(l{2pt}r{2pt}){9-14} \cmidrule(l{2pt}r{2pt}){15-16} \cmidrule(l{2pt}r{2pt}){17-18} \cmidrule(l{2pt}r{2pt}){19-21}
DIW & DIW & \quantVal{0.25} & \quantVal{0.10} & \quantVal{0.76} & \quantVal{0.62} & \quantVal{0.88} & \quantVal{0.96} & \quantVal{0.25} & \quantVal{0.10} & \quantVal{1.00} & \quantVal{0.61} & \quantVal{0.86} & \quantVal{0.95} & \quantVal{4.55} & \quantVal{41.46} & \quantVal{10.00} & \quantVal{10.00} & \quantVal{81.17} & \quantVal{8.76} & \quantVal{10.08}
\\
DIW & DIW + NYU & \quantVal{0.19} & \quantVal{0.08} & \quantVal{0.60} & \quantVal{0.73} & \quantVal{0.93} & \quantVal{0.98} & \quantVal{0.19} & \quantVal{0.08} & \quantVal{0.80} & \quantVal{0.72} & \quantVal{0.91} & \quantVal{0.97} & \quantVal{6.16} & \quantVal{30.30} & \quantVal{7.93} & \quantVal{9.41} & \quantVal{85.68} & \quantVal{7.25} & \quantVal{7.07}
\\
DeepLens & iPhone & \quantVal{0.27} & \quantVal{0.10} & \quantVal{0.82} & \quantVal{0.58} & \quantVal{0.86} & \quantVal{0.95} & \quantVal{0.26} & \quantVal{0.09} & \quantVal{1.00} & \quantVal{0.61} & \quantVal{0.86} & \quantVal{0.96} & \quantVal{7.20} & \quantVal{43.33} & \quantVal{7.48} & \quantVal{9.72} & \quantVal{80.77} & \quantVal{8.59} & \quantVal{10.64}
\\
MegaDepth & Mega & \quantVal{0.24} & \quantVal{0.09} & \quantVal{0.72} & \quantVal{0.63} & \quantVal{0.88} & \quantVal{0.96} & \quantVal{0.23} & \quantVal{0.09} & \quantVal{0.83} & \quantVal{0.67} & \quantVal{0.89} & \quantVal{0.96} & \quantVal{7.62} & \quantVal{35.51} & \quantVal{5.40} & \quantVal{8.61} & \quantVal{83.11} & \quantVal{9.05} & \quantVal{7.84}
\\
MegaDepth & Mega + DIW & \quantVal{0.21} & \quantVal{0.08} & \quantVal{0.65} & \quantVal{0.68} & \quantVal{0.91} & \quantVal{0.97} & \quantVal{0.20} & \quantVal{0.08} & \quantVal{0.78} & \quantVal{0.70} & \quantVal{0.91} & \quantVal{0.97} & \quantVal{7.04} & \quantVal{33.03} & \quantVal{4.09} & \quantVal{8.28} & \quantVal{83.74} & \quantVal{8.75} & \quantVal{7.51}
\\
Ours & Mega + NYU + Ours & \quantBes{0.08} & \quantBes{0.03} & \quantBes{0.30} & \quantBes{0.94} & \quantBes{0.99} & \quantBes{1.00} & \quantBes{0.10} & \quantBes{0.04} & \quantBes{0.47} & \quantBes{0.90} & \quantBes{0.97} & \quantBes{0.99} & \quantBes{2.17} & \quantVal{10.25} & \quantVal{2.40} & \quantVal{5.80} & \quantVal{93.48} & \quantVal{2.84} & \quantBes{3.68}
\\
Ours + Refinement & Mega + NYU + Ours & \quantBes{0.08} & \quantBes{0.03} & \quantBes{0.30} & \quantBes{0.94} & \quantBes{0.99} & \quantBes{1.00} & \quantBes{0.10} & \quantBes{0.04} & \quantBes{0.47} & \quantBes{0.90} & \quantBes{0.97} & \quantBes{0.99} & \quantVal{2.19} & \quantBes{10.24} & \quantBes{2.02} & \quantBes{5.44} & \quantBes{93.49} & \quantBes{2.83} & \quantBes{3.68}
\\ \midrule
Ours w/ DIW arch & Mega + NYU + Ours & \quantVal{0.18} & \quantVal{0.07} & \quantVal{0.56} & \quantVal{0.76} & \quantVal{0.94} & \quantVal{0.98} & \quantVal{0.15} & \quantVal{0.06} & \quantVal{0.62} & \quantVal{0.80} & \quantVal{0.95} & \quantBes{0.99} & \quantVal{6.31} & \quantVal{19.49} & \quantVal{3.12} & \quantVal{8.04} & \quantVal{89.10} & \quantVal{5.68} & \quantVal{5.22}
\\
Ours w/o our data & Mega + NYU & \quantVal{0.10} & \quantVal{0.04} & \quantVal{0.36} & \quantVal{0.90} & \quantVal{0.98} & \quantVal{0.99} & \quantVal{0.12} & \quantVal{0.05} & \quantVal{0.56} & \quantVal{0.88} & \quantBes{0.97} & \quantBes{0.99} & \quantVal{3.67} & \quantVal{16.03} & \quantVal{2.82} & \quantVal{6.30} & \quantVal{92.41} & \quantVal{3.46} & \quantVal{4.13}
		\vspace{0.04cm} \\ \bottomrule
	\end{tabularx}\vspace{0.2cm}
	\captionof{table}{Depth prediction quality. Our method compares favorably to state-of-the-art depth prediction methods in all depth quality metrics.}\vspace{-0.5cm}
	\label{tab:depthpred}
\end{figure*}

\begin{figure*}\centering
    \setlength{\tabcolsep}{0.05cm}
    \setlength{\itemwidth}{4.05cm}
    \begin{tabular}{cccc}
            \includegraphics[height=\itemwidth]{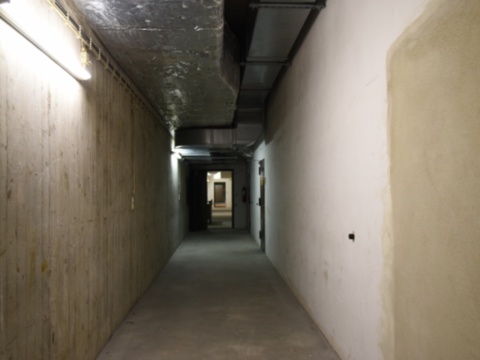}
        &
            \animategraphics[width=\itemwidth, autoplay, loop, final, nomouse, method=widget]{10}{graphics/threedim/1-lijun/}{00000}{00014}
        &
            \animategraphics[width=\itemwidth, autoplay, loop, final, nomouse, method=widget]{10}{graphics/threedim/1-megadepth/}{00000}{00014}
        &
            \animategraphics[width=\itemwidth, autoplay, loop, final, nomouse, method=widget]{10}{graphics/threedim/1-ours/}{00000}{00014}
        \\
            \footnotesize Input image.
        &
            \footnotesize Rendered depth of DeepLens.
        &
            \footnotesize Rendered depth of MegaDepth.
        &
            \footnotesize Rendering from our depth.
        \\
    \end{tabular}\vspace{-0.2cm}
	\caption{Depth-based scene rendering. Compared to off-the-shelf methods, our depth prediction pipeline often better preserves the scene geometry.}\vspace{-0.2cm}
	\label{fig:threedim}
\end{figure*}

\subsection{Automatic Mode Evaluation}
\label{sec:autoeval}

As discussed in Section~\ref{sec:automode}, our system provides an automatic mode that requires no user interaction. We investigate the effectiveness of our method in generating 3D Ken Burns effects from the input images automatically. In this experiment, we collect images from Flickr using different keywords, including ``indoor'', ``landscape'', ``outdoor'', and ``portrait'' to cover images of different scene types. We collect 12 images in total, with three images with different level of scene complexity in each category. We then use our automatic mode to generate one result for each image. For comparison, for each of our 3D Ken Burns effect result, we also generate a 2D Ken Burns effect result corresponding to the same camera path (\textit{i.e.} the same start- and end-view cropping windows).

We evaluate the quality of our results with a subjective human evaluation procedure. We recruit 21 participants to subjectively compare the quality of our 3D Ken Burns synthesis results and the 2D counterparts. Each participant performs 12 comparison sessions corresponding to our 12 test images. Each session consists of a pair-wise comparison test presenting both the 3D and 2D Ken Burns synthesis results from an image in our test set. The participant is then asked to determine the result with better quality in terms of both 3D perception and overall visual quality.

Figure~\ref{fig:user-twodee} shows average user preference percentage for our 3D Ken Burns effect results and those from the baseline 2D version for images in each category. The result indicates that our 3D Ken Burns synthesis results are preferred by the users in a majority of cases, which demonstrates the usefulness and effectiveness of our system. Please refer to our supplementary video for more visual examples of the comparison. Figure~\ref{fig:twodee} shows two examples comparing our generated 3D Ken Burns effect with the 2D version resulting from the same start- and end-view cropping windows. The 2D results show a typical zooming effect with no parallax. Our results, on the other hand, contain realistic motion parallax with strong depth perception, leading to a much more desirable effect.

\begin{figure*}\centering
    \setlength{\tabcolsep}{0.05cm}
    \setlength{\itemwidth}{5.89cm}
    \begin{tabular}{ccc}
            \begin{tikzpicture}
                \node [anchor=south west, inner sep=0.0cm] (image) at (0,0) {
                    \includegraphics[width=\itemwidth, trim={0.0cm 0.0cm 0.0cm 0.5cm}, clip]{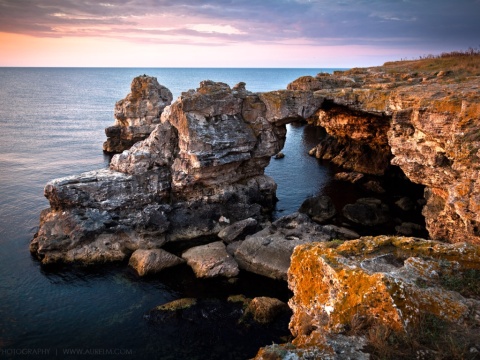}
                };
                \begin{scope}[x={(image.south east)},y={(image.north west)}]
                    \node [anchor=south west, fill=white, inner sep=0.05cm] at (0.02,0.025) {\tiny by Aurel Manea};
                \end{scope}
            \end{tikzpicture}
        &
            \animategraphics[width=\itemwidth, trim={0.0cm 0.0cm 0.0cm 0.5cm}, autoplay, palindrome, final, nomouse, method=widget, poster=last]{10}{graphics/artisan/I3IXnZrV1-k-12-artist/}{00000}{00014}
        &
            \animategraphics[width=\itemwidth, trim={0.0cm 0.0cm 0.0cm 0.5cm}, autoplay, palindrome, final, nomouse, method=widget, poster=last]{10}{graphics/artisan/I3IXnZrV1-k-12-ours/}{00000}{00014}
        \\
            \footnotesize Input image.
        &
            \footnotesize 3D Ken Burns from a professional artist.
        &
            \footnotesize 3D Ken Burns from our system.
        \\
    \end{tabular}\vspace{-0.2cm}
	\caption{Example result comparing the 3D Ken Burns effect created by a professional artist with our automatic 3D Ken Burns synthesis.}\vspace{-0.2cm}
	\label{fig:artisan}
\end{figure*}

\begin{figure}\centering
    \includegraphics[]{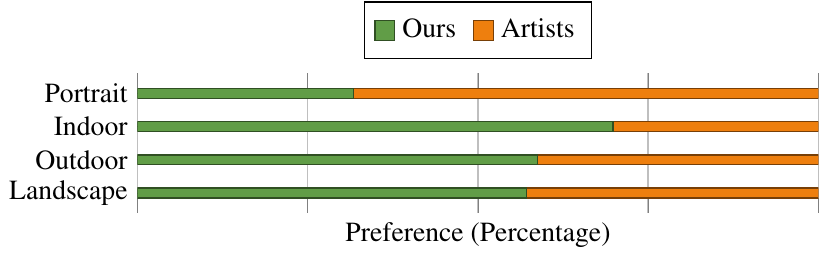}\vspace{-0.2cm}
	\caption{Results from a subjective user study comparing our 3D Ken Burns synthesis to results from artists, indicating no clear preference.}\vspace{-0.2cm}
	\label{fig:user-artisan}
\end{figure}

\subsection{Depth Prediction Quality}

We now evaluate the effectiveness of our depth prediction module. We compare our depth prediction results with those from three state-of-the-art monocular depth prediction methods, including MegaDepth~\cite{Li_CVPR_2018}, DeepLens~\cite{Wang_TOG_2018}, and DIW~\cite{Chen_NIPS_2016}. For each method, we use the publicly available implementations provided by the authors. We evaluate the depth prediction quality using two public benchmarks on single-image depth estimation. We report the performance of MegaDepth, DeepLens, and DIW with their models trained on their proposed datasets. To address the scale-ambiguity of depth estimation, we scale and shift each depth prediction to minimize the absolute error between it and the ground truth.

\textit{NYU~v2.} \cite{Silberman_ECCV_2012} created one of the most well-known a benchmarks and datasets for single-image-depth estimation, consisting of 464 indoor scenes. Each scene contains aligned RGB and depth images, acquired from a Microsoft Kinect sensor. Following previous works on single-image depth estimation~\cite{Chen_NIPS_2016, Qi_CVPR_2018, Zoran_ICCV_2015}, we use the standard training-testing split and evaluate our method on the 654 image-depth pairs from the testing set.

\textit{iBims-1.} Recently \cite{Koch_ARXIV_2018} introduced a new benchmark aiming for a more holistic evaluation of the depth prediction quality. This benchmark consists of 100 images with high-quality ground-truth depth maps. These images cover a wide variety of indoor scenes and the benchmark provides a comprehensive set of quality metrics to quantify different desired properties of a well-predicted depth map such as depth boundary quality, planarity, depth consistency, and absolute distance accuracy.

Table~\ref{tab:depthpred} (top) compares the depth prediction quality of different methods according to various quantitative metrics defined by each benchmark. Our method compares favorably to state-of-the-art depth prediction methods in all depth quality metrics. In addition, the result demonstrates that our depth prediction pipeline improves significantly over off-the-shelf methods in terms of the Planarity Error (PE) and Depth Boundary Error (DBE) metrics on the iBims-1 benchmark. Those metrics are particularly designed to assess the quality in planarity and depth boundary preservation, respectively, which are particularly important for our synthesis task.

Table~\ref{tab:depthpred} (bottom) lists two additional variations of our approach to better analyze the effect of our depth estimation network as well as our training dataset. Specifically, we supervised the network architecture from DIW~\cite{Chen_NIPS_2016} with all available training data to compare this architecture to ours. Furthermore, we supervised our depth estimation network only on the training data from MegaDepth and NYU~v2 without incorporating our computer-generated dataset. Both variants lead to significantly worse depth quality metrics in the benchmark, which exemplifies the importance of all individual components of our proposed approach. Interestingly, both variants compare favorably to state-of-the-art depth prediction models.

Figure~\ref{fig:threedim} compares the three-dimensional renderings with respect to different depth prediction results. We can observe better preservation of the scene structure such as the planarity in our result compared to off-the-shelf depth prediction methods.

\subsection{Discussion}

Our previous experiment in Section~\ref{sec:autoeval} shows that users prefer our 3D Ken Burns effects in favor of the traditional 2D Ken Burns technique. It is also interesting to investigate how the effects created by our method compare to the ones made by skilled professional artists through laborious manual processing. 

We conduct an additional subjective evaluation test. For each of the 12 artist-generated 3D Ken Burns results that we collected in Section~\ref{sec:usability}, we use our system to create similar 3D Ken Burns effects using the corresponding input image. For each of the 12 test examples, we thus have a reference result generated by an artist and our result created by our proposed system. Please see Figure~\ref{fig:artisan} for an example. We follow the same procedure as in Section~\ref{sec:autoeval}. We ask the same set of 21 participants to perform 12 additional pair-wise comparison tests, comparing the results created by our system with the original artist-generated ones. 

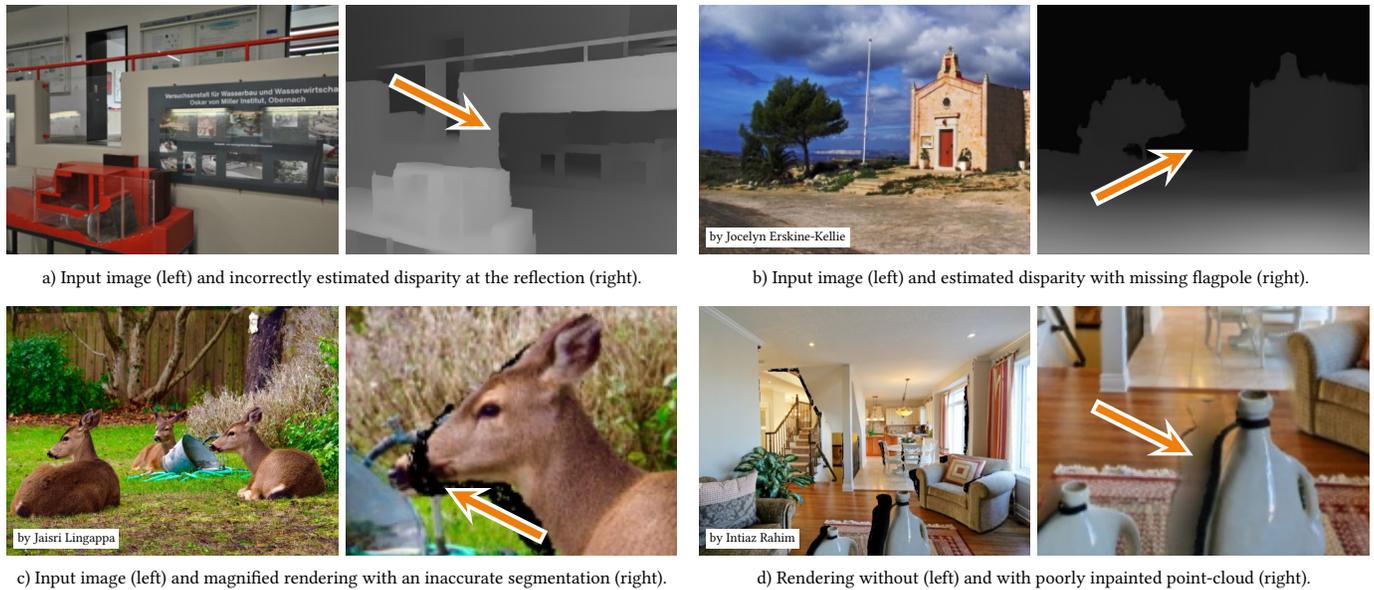
\begin{figure*}\centering
    \setlength{\tabcolsep}{0.05cm}
    \setlength{\itemwidth}{4.4cm}
    \begin{tabular}{cccccc}
            \includegraphics[width=\itemwidth]{{{graphics/limitations/corridor_07.png-input}}}
        &
            \begin{tikzpicture}
                \definecolor{arrowcolor}{RGB}{238,127,14}
                \node [anchor=south west, inner sep=0.0cm] (image) at (0,0) {
                    \includegraphics[width=\itemwidth]{{{graphics/limitations/corridor_07.png-disparity}}}
                };
                \begin{scope}[x={(image.south east)},y={(image.north west)}]
                    \draw [double arrow=0.2cm with white and arrowcolor] (0.44-0.3,0.5+0.2) -- (0.44,0.5);
                \end{scope}
            \end{tikzpicture}
        &&&
            \begin{tikzpicture}
                \node [anchor=south west, inner sep=0.0cm] (image) at (0,0) {
                    \includegraphics[width=\itemwidth, trim={0.0cm 0.0cm 7.5cm 5.625cm}, clip]{{{graphics/limitations/11685197874_ecda9c1d68_b.jpg-input}}}
                };
                \begin{scope}[x={(image.south east)},y={(image.north west)}]
                    \node [anchor=south west, fill=white, inner sep=0.05cm] at (0.02,0.025) {\tiny by Jocelyn Erskine-Kellie};
                \end{scope}
            \end{tikzpicture}
        &
            \begin{tikzpicture}
                \definecolor{arrowcolor}{RGB}{238,127,14}
                \node [anchor=south west, inner sep=0.0cm] (image) at (0,0) {
                    \includegraphics[width=\itemwidth, trim={0.0cm 0.0cm 7.5cm 5.625cm}, clip]{{{graphics/limitations/11685197874_ecda9c1d68_b.jpg-disparity}}}
                };
                \begin{scope}[x={(image.south east)},y={(image.north west)}]
                    \draw [double arrow=0.2cm with white and arrowcolor] (0.47-0.3,0.42-0.2) -- (0.47,0.42);
                \end{scope}
            \end{tikzpicture}
        \\
            \multicolumn{2}{c}{\footnotesize a) Input image (left) and incorrectly estimated disparity at the reflection (right).}
        &&&
            \multicolumn{2}{c}{\footnotesize b) Input image (left) and estimated disparity with missing flagpole (right).}
        \vspace{0.2cm} \\
            \begin{tikzpicture}
                \node [anchor=south west, inner sep=0.0cm] (image) at (0,0) {
                    \includegraphics[width=\itemwidth]{{{graphics/limitations/27624269639_33b396843d_o.jpg-input}}}
                };
                \begin{scope}[x={(image.south east)},y={(image.north west)}]
                    \node [anchor=south west, fill=white, inner sep=0.05cm] at (0.02,0.025) {\tiny by Jaisri Lingappa};
                \end{scope}
            \end{tikzpicture}
        &
            \begin{tikzpicture}
                \definecolor{arrowcolor}{RGB}{238,127,14}
                \node [anchor=south west, inner sep=0.0cm] (image) at (0,0) {
                    \includegraphics[width=\itemwidth, trim={11.0cm 5.0cm 1.25cm 4.185cm}, clip]{{{graphics/limitations/27624269639_33b396843d_o.jpg-raw}}}
                };
                \begin{scope}[x={(image.south east)},y={(image.north west)}]
                    \draw [double arrow=0.2cm with white and arrowcolor] (0.3+0.3,0.27-0.2) -- (0.3,0.27);
                \end{scope}
            \end{tikzpicture}
        &&&
            \begin{tikzpicture}
                \node [anchor=south west, inner sep=0.0cm] (image) at (0,0) {
                    \includegraphics[width=\itemwidth]{{{graphics/limitations/3407755658_1f1a834b61_o.jpg-raw}}}
                };
                \begin{scope}[x={(image.south east)},y={(image.north west)}]
                    \node [anchor=south west, fill=white, inner sep=0.05cm] at (0.02,0.025) {\tiny by Intiaz Rahim};
                \end{scope}
            \end{tikzpicture}
        &
            \begin{tikzpicture}
                \definecolor{arrowcolor}{RGB}{238,127,14}
                \node [anchor=south west, inner sep=0.0cm] (image) at (0,0) {
                    \includegraphics[width=\itemwidth, trim={6.5cm 0.0cm 4.5cm 8.245cm}, clip]{{{graphics/limitations/3407755658_1f1a834b61_o.jpg-ours}}}
                };
                \begin{scope}[x={(image.south east)},y={(image.north west)}]
                    \draw [double arrow=0.2cm with white and arrowcolor] (0.47-0.3,0.4+0.2) -- (0.47,0.4);
                \end{scope}
            \end{tikzpicture}
        \\
            \multicolumn{2}{c}{\footnotesize c) Input image (left) and magnified rendering with an inaccurate segmentation (right).}
        &&&
            \multicolumn{2}{c}{\footnotesize d) Rendering without (left) and with poorly inpainted point-cloud (right).}
        \\
    \end{tabular}\vspace{-0.2cm}
	\caption{Examples of various commonly occurring issues with our proposed approach. Please see the limitations section for further details.}\vspace{-0.2cm}
	\label{fig:limitations}
\end{figure*}

Figure~\ref{fig:user-artisan} shows user preference percentage averaged over test cases in each category. Interestingly, our results are rated on-par with the ones from professional artists. Looking closely into each individual category, we observe that our results are slightly preferred compared to the artist's results in the indoor category. These scenes typically have a complicated depth distribution with many objects, which makes it extremely tedious to manually achieve the 3D Ken Burns effect. Our method can rely on a good depth prediction to handle those complicated scenes. The artist-created results, however, are more preferred in the portrait category. Looking into the results, we observe that portrait images often have simpler scene layouts which makes it easier to manually achieve good results. More importantly, we found that artists often intentionally exaggerate the parallax effect in portrait photos to make the effect much more dramatic to an extent that is not possible with physically-correct depth. This artistic emphasis is often preferred by viewers. Our method is limited by the parallax enabled by our depth prediction which is trained to match physically-correct depth and thus is not able to generate such dramatic effects.

We hope that our geometric- and semantic-aware depth prediction framework provides useful insights for future research in developing a more effective depth prediction tailored to view synthesis tasks. We would in this regard like to emphasize that the 3D Ken Burns effect is an artistic effect. In certain scenarios, view synthesis results generated from a physically correct scene prediction may not be optimal in delivering the desired artistic impression. Allowing such artistic manipulation in the 3D Ken Burns effect synthesis is an interesting direction to extend our work in the future.

\subsection{Limitations}

While our method can generate a plausible 3D Ken Burns effect for images of different scene types, the results are not always perfect as shown in Figure~\ref{fig:limitations}. Single image depth estimation is highly challenging and our semantic-aware depth estimation network is not infallible. While our method can produce depth estimates subject to little or no distortion, we found that our results may still fail to predict accurate depth maps for challenging cases such as reflective surfaces (the reflection on the glossy poster in Fig.~\ref{fig:limitations}~(a)) or thin structures (the flagpole in Fig.~\ref{fig:limitations}~(b)). Object segmentation is challenging as well and the salient depth adjustment may fail due to erroneous masks. While our depth upsamling module can perform boundary-aware refinement to account for some mask inaccuracies, our result is affected when the error in the segmentation mask is significantly large. In Fig.~\ref{fig:limitations}~(c), the nose of the deer is cut off due to Mask R-CNN providing an inaccurate segmentation. Finally, we note that while our joint color- and depth-inpainting is an intuitive approach to extend the estimated scene geometry, it has only been supervised on our synthetic data and thus may sometimes generate artifacts when the input differs too much from the training data. In Fig.~\ref{fig:limitations}~(d), the inpainting result lacks texture and is darker than expected. Training the color- and depth-inpainting model with real images and leveraging an adversarial supervision regime and a more sophisticated architecture, like one that uses partial convolutions, is an interesting direction to explore in future work.

\section{Conclusion}
\label{sec:conclusion}

In this paper, we developed a complete framework to produce the 3D Ken Burns effect from a single input image. Our method consists of a depth prediction model which predicts scene depth from the input image and a context-aware depth-based view synthesis model to generate the video results. To this end, we presented a semantically-guided training strategy along with high-quality synthetic data to train our depth prediction network. We couple its prediction with a semantics-based depth adjustment and a boundary-focused depth refinement process to enable an effective depth prediction for view synthesis. We subsequently proposed a depth-based synthesis model that jointly predicts the image and the depth map at the target view using a context-aware view synthesis framework. Using our synthesis model, the extreme views of the camera path are synthesized from the input image and the predicted depth map, which can be used to efficiently synthesize all intermediate views of the target video, resulting in the final 3D Ken Burns effect. Experiments with a wide variety of image content show that our method enables realistic synthesis results. Our study shows that our system enables users to achieve better results while requiring little effort compared to existing solutions for the 3D Ken Burns effect creation.

\begin{acks}
    This work was done while Simon was interning at Adobe Research. We would like to thank Tobias Koch for his help with the iBims-1 benchmark. We are grateful for being allowed to use footage from Ian D. Keating (Figure~\ref{fig:teaser}, top), Kirk Lougheed (Figure~\ref{fig:teaser}, bottom), Leif Skandsen (Figure~\ref{fig:depthestim}, top), Oliver Wang (Figure~\ref{fig:depthestim}, bottom), Ben Abel (Figure~\ref{fig:overview-depth}, \ref{fig:steps}, \ref{fig:rendering}, \ref{fig:inpainting}, \ref{fig:overview-synthesis}), Aurel Manea (Figure~\ref{fig:artisan}), Jocelyn Erskine-Kellie (Figure~\ref{fig:limitations}, top right), Jaisri Lingappa (Figure~\ref{fig:limitations}, bottom left), and Intiaz Rahim (Figure~\ref{fig:limitations}, bottom right).
\end{acks}

{\small
\bibliographystyle{ACM-Reference-Format}
\bibliography{main}
}

\end{document}